\documentclass[10pt,journal,compsoc]{IEEEtran}
\usepackage[nocompress]{cite}

\usepackage[margin=0pt,font=small,labelfont=bf,labelsep=endash,tableposition=top]{caption}

\usepackage{amsmath}
\usepackage{array}

\usepackage[caption=false,font=footnotesize,labelfont=sf,textfont=sf]{subfig}
\usepackage{fixltx2e}

\usepackage{url}
\usepackage{graphicx}
\usepackage{amsmath}
\usepackage{amssymb}

\usepackage{epsfig}
\usepackage{epstopdf}
\usepackage{makecell,multirow,diagbox}
\usepackage{color}
\usepackage{soul}
\usepackage{url}
\usepackage{array}
\usepackage{dsfont}
\usepackage[utf8]{inputenc}
\usepackage{xcolor}
\usepackage{hyperref}
\usepackage{helvet}
\usepackage{courier}
\usepackage{bm}
\usepackage{bbm}
\usepackage{wrapfig}
\usepackage{picinpar}
\usepackage{arydshln}
\usepackage{epsfig}
\usepackage{epstopdf}
\usepackage{makecell}
\usepackage{multicol}
\usepackage{booktabs}
\usepackage{subcaption}

\newlength\savedwidth

\newlength\savewidth

\usepackage[vlined,ruled,linesnumbered]{algorithm2e}
\usepackage{cite}
\usepackage{balance}
\usepackage{amssymb}

\usepackage{mathtools,xspace}

\usepackage[utf8]{inputenc}
\usepackage{xcolor}

\def\etal{{\it et al.}\xspace}

\hypersetup{
    colorlinks=true,
    breaklinks=true,
    urlcolor= blue,
    linkcolor= red,
    bookmarksopen=false,
    filecolor=black,
    citecolor=blue,
    linkbordercolor=blue
}

\AtBeginDocument{\hypersetup{pdfborder={0 0 1}}}

\begin{document}

\title{
VimTS: A Unified Video and Image Text Spotter for Enhancing the Cross-domain Generalization
}

\author{
Yuliang Liu, IEEE Member, 
Mingxin Huang,
Hao Yan,
Linger Deng,
Weijia Wu,
Hao Lu, \\
Chunhua Shen, 
Lianwen Jin*, IEEE Member,
Xiang Bai, IEEE Senior Member

\thanks{ 
Y. Liu, H. Yan, L. Deng, H. Lu, and X. Bai are with School of Artificial Intelligence and Automation, Huazhong University of Science and Technology, Wuhan, 430074, China.
M. Huang and L. Jin are with South China University of Technology, Guangzhou, 510000, China (email: eelwjin@scut.edu.cn).
W. Wu and C. Shen are with Zhejiang University, Zhejiang, 310058, China. \\

Corresponding author: L. Jin.
}
}

\IEEEtitleabstractindextext{%
\begin{abstract}

Text spotting, a task involving the extraction of textual information from image or video sequences, faces challenges in cross-domain adaption, such as image-to-image and image-to-video generalization. In this paper, we introduce a new method, termed VimTS, which enhances the generalization ability of the model by achieving better synergy among different tasks. Typically, we propose a Prompt Queries Generation Module and a Tasks-aware Adapter to effectively convert the original single-task model into a multi-task model suitable for both image and video scenarios with minimal additional parameters. The Prompt Queries Generation Module facilitates explicit interaction between different tasks, while the Tasks-aware Adapter helps the model dynamically learn suitable features for each task. Additionally, to further enable the model to learn temporal information at a lower cost, we propose a synthetic video text dataset (VTD-368k) by leveraging the Content Deformation Fields (CoDeF) algorithm. Notably, our method outperforms the state-of-the-art method by an average of $2.6\%$ in six cross-domain benchmarks such as TT-to-IC15, CTW1500-to-TT, and TT-to-CTW1500. For video-level cross-domain adaption, our method even surpasses the previous end-to-end video spotting method in ICDAR2015 video and DSText v2 by an average of $5.5\%$ on the MOTA metric, using only image-level data.
We further demonstrate that existing Large Multimodal Models exhibit limitations in generating cross-domain scene text spotting, in contrast to our VimTS model which requires significantly fewer parameters and data.
The code and datasets will be made available at the \url{https://VimTextSpotter.github.io}.
\end{abstract}

}

\maketitle

\IEEEdisplaynontitleabstractindextext
\IEEEpeerreviewmaketitle

\IEEEraisesectionheading{\section{Introduction}\label{sec:introduction}}

\IEEEPARstart{T}{ext} spotting has garnered significant attention in recent years, driven by its potential applications such as automated subtitles in dynamic scenes, real-time reading of road signs for self-driving vehicles, and instant translation for international broadcasts. From two-stage methods~\cite{liu2018fots,liao2020masktext,liu2021abcnetv2} to Transformer-based method~\cite{zhang2022text,kittenplon2022towards,peng2022spts}, the relationship between detection and recognition is getting closer and has achieved significant advancements. 

\begin{figure}[t!]
    \centering
    \includegraphics[width=\linewidth]{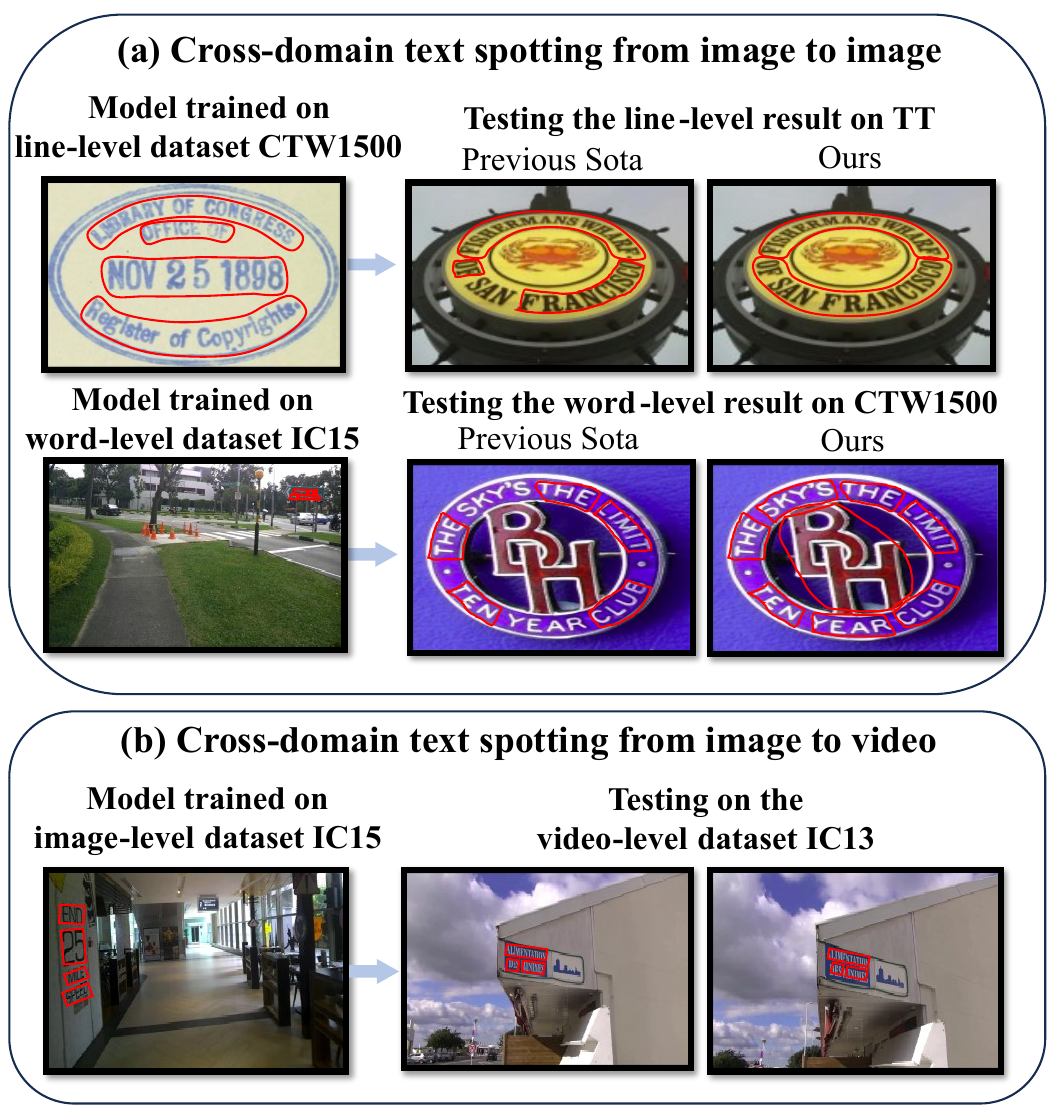}
    \caption{Fig. (a) and (b) are two types of cross-domain text spotting, including image-to-image and image-to-video. TT represents the TotalText. 
    IC15 represents the ICDAR2015~\cite{karatzas2015icdar}. IC13 represent the ICDAR2013 video~\cite{karatzas2013icdar}. TT represents the TotalText~\cite{ch2019total}.
    }
    \label{fig:intro}
\end{figure}

While text spotting has shown promising progress, addressing cross-domain text spotting remains a significant challenge that requires further exploration~\cite{yu2023turningts}. In this paper, the cross-domain text spotting encompasses image-level and video-level scenarios, as shown in Fig.~\ref{fig:intro} (a) and (b). Image-level cross-domain text spotting involves the migration between images, while video-level cross-domain text spotting represents the generalization from images to video scenes.

1) In image-level cross-domain text spotting, the challenge lies in the varying styles, fonts, and backgrounds across different benchmarks, requiring models to generalize well beyond their training data. When the training set and the test set belong to different benchmarks, the model usually performs poorly, as demonstrated in ~\cite{yu2023turningts}. Furthermore, the differences in annotation formats across various datasets bring further challenges. For instance, TotalText~\cite{ch2019total} and ICDAR2015~\cite{karatzas2015icdar} utilize word-level annotations, while CTW1500~\cite{liu2019curved} employs line-level annotations. Integrating datasets with distinct annotation formats simultaneously proves challenging. This disparity not only complicates the training process but also hampers model performance, as the model struggles to adapt to divergent annotation schemes.

    \begin{figure*}[t!]
        \centering
        \includegraphics[width=\linewidth]{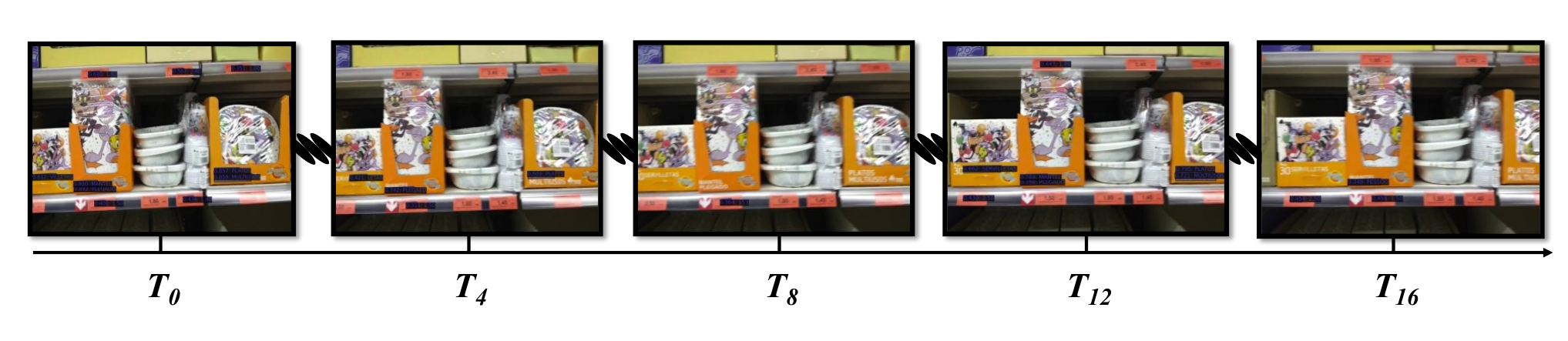}
        \caption{Applying static text spotting methods (TESTR, results shown in the image) to videos, even those with minimal motion, leads to poor performance in both bounding box recall and recognition accuracy.}
        \label{fig:intro2}
    \end{figure*}

2) For video-level cross-domain text spotting, intuitively, it might appear simple and efficient to train models on static images for text spotting and then apply these models frame-by-frame to videos. However, as illustrated in Fig.~\ref{fig:intro2}, we found that models tailored for static images frequently underperform in video contexts, resulting in a significant number of missed detections. Videos incorporating dynamic factors like occlusions, swift changes in scenes, and temporal relationships between frames, the involvement of additional video text tracking tasks diminishes the effectiveness and accuracy of text spotters designed for still images. Therefore, exploring methods to effectively adapt text spotters for video contexts is crucial. When we further explore the video text recognition, we find that the video text data is relatively lacking, due to the expensive annotation cost. Unlike text detection, text spotting normally requires significantly more training data to achieve acceptable text recognition performance. A solution is using the optical flow estimation for data synthesis~\cite{10219970}. However, this method presents several challenges, including distortion, labeling errors, and a bias towards static objects. Additionally, this method~\cite{10219970} does not provide open-source synthetic data for communal utilization. There is a need for open-sourced large-scale synthetic data in video text spotting, such as SynthText~\cite{gupta2016synthetic} for static images. 

In this paper, we present VimTS, a unified framework capable of performing both word-level and line-level text spotting, as well as video-level text spotting. Through the unified framework, the model is allowed to leverage the synergy between different tasks. 
When data for one task is lacking in a certain scenario, we improve cross-domain performance by jointly optimizing tasks across different scenarios. 
Specifically, to achieve synergy between different tasks, we develop two forms of interaction. Firstly, inspired by~\cite{huang2023estextspotter}, VimTS employs a unique decomposition of conventional queries into specific queries for text detection, recognition, and tracking, which allows queries for different tasks to conduct explicit interaction between each other. 
To achieve synergy among hierarchical text spotting tasks, we propose a Prompt Queries Generation Module, which performs the interaction between word-level, line-level, and video-level text spotting.
To efficiently learn these interactive and discriminative features, we introduce a Tasks-aware Adapter to dynamically select suitable features for different tasks. The PQGM with the Task-aware Adapter can effectively transform the original model into a multi-task model for both image and video scenarios with minimal additional parameters. 
Furthermore, we utilize the CoDeF~\cite{ouyang2023codef} to facilitate the achievement of realistic and stable text flow propagation for constructing a synthetic video text dataset (VTD-368k). This dataset augments VimTS's learning of temporal information, while image-level data help VimTS learn detection and recognition capabilities in real-world scenarios, thus efficiently adapting the model pre-trained on image to video.

The advantages of our method can be summarized as follows:
\begin{itemize}
\item 
We introduce a new framework, termed VimTS, designed to leverage the synergy between various tasks and scenarios, thereby enhancing the generalization ability in text spotting.

\item 
We propose a Prompt Queries Generation Module and a Task-aware Adapter, which combines hierarchical text spotting tasks into a unified framework, requiring only minimal additional parameters. The training of task-aware adapter only requires 3.0\% of parameters compared to fine-tuning the entire model.

\item 
We propose VTD-368k, a large-scale synthetic video text dataset, derived from high-quality open-source videos, designed to improve text spotting models. By utilizing the CoDeF to facilitate the achievement of realistic and stable text flow propagation, the synthetic video data is more consistent with the text between frames.

\item 
In image-level cross-domain text spotting, our method demonstrates superior performance compared to the state-of-the-art method across six benchmarks, with an average improvement of $2.6\%$ in terms of Hmean. Furthermore, in video-level cross-domain text spotting, our method surpasses the previous end-to-end approach in both ICDAR2015 video and DSText v2 by an average of $5.5\%$ on MOTA metric, achieved solely through the utilization of image-level data. Additionally, by using the video data, such as VTD-368k or ICDAR2015 video, our method can achieve further improvement.
\end{itemize}

\section{Related Work}
\label{sec:related-works}

\subsection{Scene Text Spotting}
Scene text spotting has seen a significant shift from conventional methodologies to more modern end-to-end learning techniques. Traditional approaches, as highlighted in previous research  \cite{jaderberg2016reading}, often faced challenges related to inference inefficiencies and error accumulation. Pioneering a more unified perspective, Li \etal~\cite{li2017towards} integrated detection and recognition into a unified framework, creating a comprehensive framework, which predominantly focused on horizontal texts.

As the field progressed, solutions for handling texts in varied orientations emerged. Techniques such as Text-Align \cite{he2018end} and RoI-Rotate \cite{liu2018fots} became instrumental in transforming oriented text features into horizontal ones. An innovative approach Mask TextSpotter \cite{liao2019mask}, leveraging character-level segmentation to recognize and process irregularly shaped texts. TextDragon~\cite{feng2019textdragon} viewed text instances as a set of segments and introduced a RoISlide to curved texts. Qin \etal~\cite{qin2019towards} proposed RoI
Masking to suppress the background noise for the recognition features. Concurrently, MaskTextSpotter v3~\cite{liao2020masktext} proposes a Segmentation Proposal Network (SPN) to generate accurate proposals for arbitrarily-shaped text. PAN++~\cite{wang2021pan++} presents a fast framework based on a fast text detector~\cite{wang2019efficient}. ABCNet~\cite{liu2020abcnet} used the parametric bezier curve to model the curved text and developed BezierAlign for rectifying curved text. ABINet++~\cite{fang2022abinet++} attempted to introduce an autonomous, bidirectional, and iterative language model~\cite{fang2021read} to the recognizer. To enhance long text recognition, ABINet++ integrates character order and content to attend to character features precisely. TESTR~\cite{zhang2022text} adopted a dual-decoder structure for detection and recognition based on the Deformable DETR~\cite{zhu2020deformable}. Other recent contributions like SwinTextSpotter~\cite{huang2022swintextspotter} endeavored to fuse detection and recognition more closely. Similarly, SPTS series~\cite{peng2022spts,liu2023spts} adopted an auto-regressive framework in which most parameters are shared between text detection and recognition. TTS~\cite{kittenplon2022towards} introduces a weakly-supervised multi-task transformer that achieves competitive performance with previous methods solely through training with text transcription. Ye \etal~\cite{ye2023deepsolo} further introduces a set of sequential queries for both detection and recognition within a single decoder.

    \begin{figure*}[t!]
        \centering
        \includegraphics[width=0.95\linewidth]{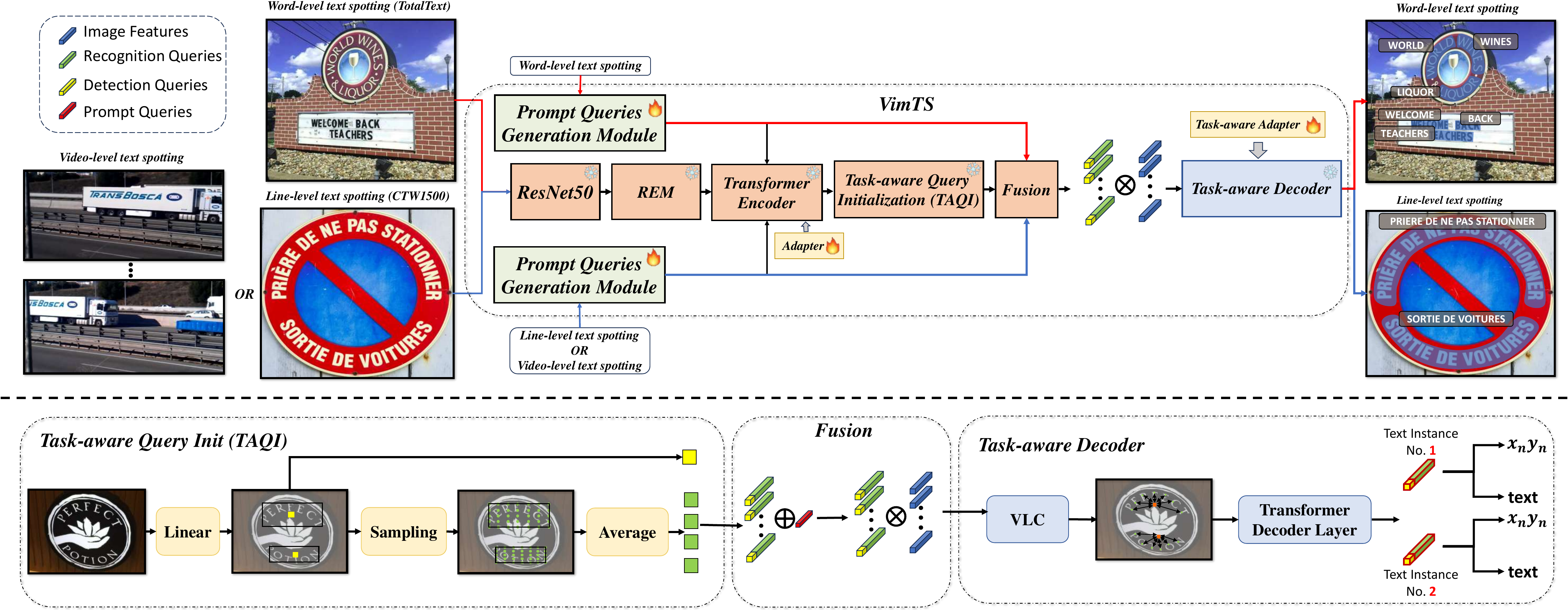}
        \caption{Overall framework of the VimTS. 
        The image features are extracted in the feature extraction process. Then, the Query Initialization is used to generate the task-aware queries including detection and recognition queries. The task-aware queries are sent to the task-aware decoder to obtain the detection and recognition results simultaneously. The red arrow means performing word-level text spotting. The blue arrow means performing line-level text spotting. After the model is pre-trained, we freeze most of its parameters and train only the Task-ware Adapter and Prompt Queries Generation Module to convert the original single-task model into a multi-task model.}
        \label{fig:overall}
    \end{figure*}

\subsection{Video Text Spotting}
Compared to static text spotting, there have been few recent advancements in video text spotting due to its challenges. Yin \etal~\cite{yin2016text} surveyed text detection, tracking, and recognition in video. Wang \etal~\cite{wang2017end} developed an end-to-end system for video text spotting, utilizing hand-crafted post-processing for tracking. Subsequently, Cheng \etal~\cite{cheng2019you} integrated a spatial-temporal video text detector with a text recommender to select high-quality text and perform recognition once. Its improved version, FREE~\cite{cheng2020free}, adopted an end-to-end trainable framework for faster and robust video text spotting. Nguyen \etal~\cite{nguyen2014video} performed multi-scale character detection and exploited temporal redundancy to remove false positives. Rong \etal~\cite{rong2014scene} proposed a framework for scene text recognition across multiple frames using feature representation and a conditional random field model. Wu \etal~\cite{wu2021bilingual} developed a dual decoder framework for detection and tracking, utilizing object features and IoU match. Inspired by the MOTR~\cite{zeng2022motr}, Wu \etal~\cite{wu2022end} unified text detection and tracking into a single decoder, treating video text spotting as a long-range temporal variation problem. To further enhance the performance of the recognizer, VLSpotter~\cite{Zu2023vlspotter} introduced a super-resolution module and language model for video text spotting.

\subsection{Domain Adaptation}
Domain adaptation, a burgeoning field of study, addresses the challenges posed by domain shift and dataset bias. It has been studied extensively in many fields~\cite {tzeng2017adversarial,you2019universal,ge2023domain}. Researchers also proposed several cross-domain approaches in OCR. For instance, Chen \etal~\cite{chen2019cross} propose a pixel-level domain adaptation block and an image-level domain adaptation block for text detection to reduce the domain shift by introducing adversarial training. Wu \etal~\cite{wu2020synthetic} attempt to transfer knowledge from synthetic data to real data for scene text detection. They employ text instance alignment to facilitate learning domain-invariant features adversarially, and utilize text self-training to reduce the impact of inaccurate pseudo-labels. 
Additionally, Chen \etal~\cite{chen2021self} develop a self-training framework aimed at automatically extracting the pseudo-labels of challenging text instances from unannotated videos or images. Yu \etal~\cite{yu2023turning} introduce a cross-modal interaction mechanism, integrating vision-language information from CLIP~\cite{radford2021learning} to improve the cross-domain performance of scene text detectors. The above methods are designed for scene text detection. 

For scene text recognition, Zhang \etal~\cite{zhang2019sequence} introduced SSDAN, a sequence-to-sequence domain adaptation network. SSDAN utilizes a gated attention similarity unit to dynamically align the distribution of source and target sequence data within a character-level feature space. GA-DAN~\cite{zhan2019ga} develops a geometry-aware domain adaptation network for both text detection and recognition. It effectively handles cross-domain shifts by simultaneously modeling changes in both geometric and appearance spaces, enabling realistic image conversion between domains with diverse characteristics. 
E2STR~\cite{zhao2024multi} presents an in-context training strategy for scene text recognition, which enables text recognizer to perform rapid adaptation across diverse scenarios without additional fine-tuning. In recent years, there have also been ways to start looking at cross-domain aspects of text spotting. Yu \etal~\cite{yu2023turningts} build on their prior work~\cite{yu2023turning} by introducing FastTCM-CR50, a new backbone for scene text detection and spotting that improves cross-domain performance. It integrates an implicit image condition and bimodal similarity matching, enabling offline CLIP computations to boost accuracy and reduce inference time.

\section{
Methodology
}
In this paper, we introduce VimTS, a unified framework designed to leverage the synergy between various tasks. Through this unified framework, this approach aims to improve the generalization ability of the text spotter. The overall architecture is illustrated in Fig.~\ref{fig:overall}. Inspired by~\cite{huang2023estextspotter}, we employ a set of task-aware queries to represent various tasks. Initially, image features are obtained via a feature extraction process comprising ResNet50, the receptive enhancement module (REM), and the Transformer encoder. Similar to~\cite{huang2023estextspotter}, these features are then used to generate task-aware queries through the Query Initialization module, which encompasses both detection and recognition queries. The Initialization of tracking queries is presented in Sec.~\ref{sec:tracking_queries}. Subsequently, these queries are fed into the task-aware decoder~\cite{huang2023estextspotter} to explicitly capture discriminative and interactive features for text detection, recognition, and tracking simultaneously. Then, a Prompt Queries Generation Module (PQGM) and a Task-aware Adapter are used to enable the interaction between the hierarchical tasks including word-level and line-level text spotting, as well as video-level text spotting. During this stage of training, most parameters are frozen. Then, multitasking features are learned by the Task-aware Adapter and PQGM. Firstly, we input the prompt into the PQGM for the task to be performed. Then, the PQGM will generate prompt queries and the prompt queries are sent to the Transformer encoder and Task-aware decoder to guide the model to complete the corresponding task. It is worth noting that our method can not only perform image-level cross-domain but also learn video-level cross-domain adaption. Detailed implementations will be provided in the following subsections.

\subsection{Feature Extraction} 
For the feature extraction, we follow the~\cite{huang2023estextspotter}. We first use a ResNet50~\cite{he2017deep} to extract the features. Then, to further enlarge the receptive field of the features from ResNet50, we use a receptive enhancement module (REM). To achieve this, REM uses a convolutional layer with a large kernel to downsample the feature map. The output of the REM and ResNet are sent into a Transformer encoder~\cite{zhu2020deformable} to further enhance the long-range dependencies across various scales.

\subsection{Query Initialization}
\label{TAQI}
For the query initialization, we follow~\cite{huang2023estextspotter} and adopt the different initialization methods for different queries. We first use a liner layer to output the coarse bounding box coordinates and probabilities. Then, the top $N$ coarse bounding box coordinates are selected based on the probabilities. Then, the recognition queries are extracted within the coarse bounding box coordinates and the detection queries are generated by using a linear layer to transform the bounding box coordinates.

\subsection{Decoder}
\label{ESTextSpotter decoder}
For the decoder, we adopt the decoder in~\cite{huang2023estextspotter} to enhance the interaction between different tasks. The decoder consists of the vision-language communication module and the Transformer decoder layer~\cite{zhu2020deformable}. The task-aware queries, including detection queries, recognition queries, and tracking queries. The tracking queries are detailed in Sec.~\ref{sec:tracking_queries}. The queries are first fed into a vision-language communication module to enable the detection and recognition queries to interact with each other. Then, we adopt intra-group self-attention and inter-group self-attention to further aggregate the features in the Transformer decoder layer. The inter-group self-attention module is utilized to perform the interaction between tracking queries. The inter-group self-attention is used to enable detection and recognition queries to interact with each other.

\begin{figure}[!t]
    \centering
    \includegraphics[width=0.75\linewidth]{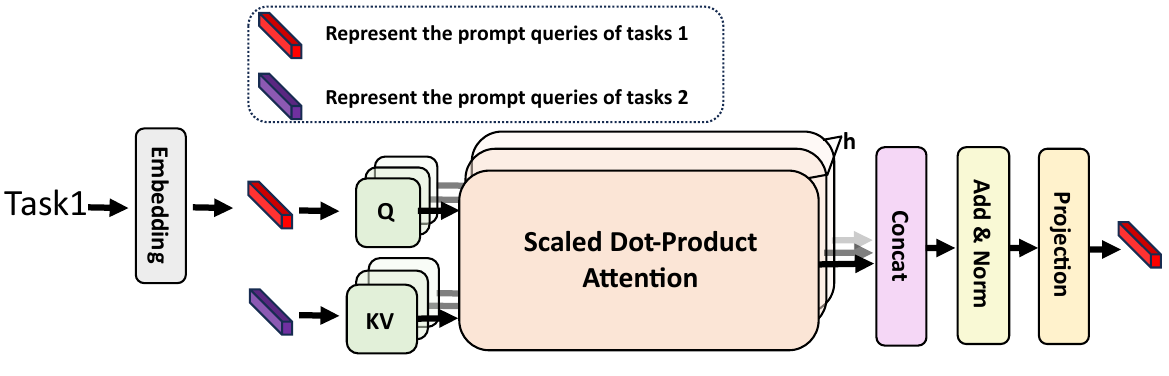}
    \caption{
        Illustration of the Prompt Queries Generation Module. Prompt queries for different tasks exchange information in the prompt queries generation module. $h$ is the number of parallel attention heads.
    }
    \label{fig:PQGM}
\end{figure}

\subsection{Prompt Queries Generation Module}
To enhance the model's ability to handle multiple tasks effectively, we propose the Prompt Queries Generation Module (PQGM) for generating prompt queries to guide its operation. The structure of this module is illustrated in Fig. \ref{fig:PQGM}. 
When inputting a prompt for a task, such as word-level text spotting, we use an embedding layer to transform the task prompt into prompt queries. These prompt queries then interact with queries from other tasks via an attention mechanism. Following the interaction, we integrate the prompt queries with the image features by directly adding them together. The combined features are then fed into the Transformer encoder for further processing and fusion. To adapt these prompt queries to learn the task-specific features, we adopt the adapter~\cite{houlsby2019parameter} in the Transformer encoder. This adapter consists of two linear projection layers and a nonlinearity activation function. These task-specific features are then utilized in Query Initialization and the Transformer decoder. To further enhance the learning of task-specific features, we further add the prompt queries to the task-aware queries from the Query Initialization.
By the prompt of the prompt queries, the model is guided toward completing the corresponding tasks more effectively.

\subsection{Task-aware Adapter}
\begin{figure}[!t]
    \centering
    \includegraphics[width=\linewidth]{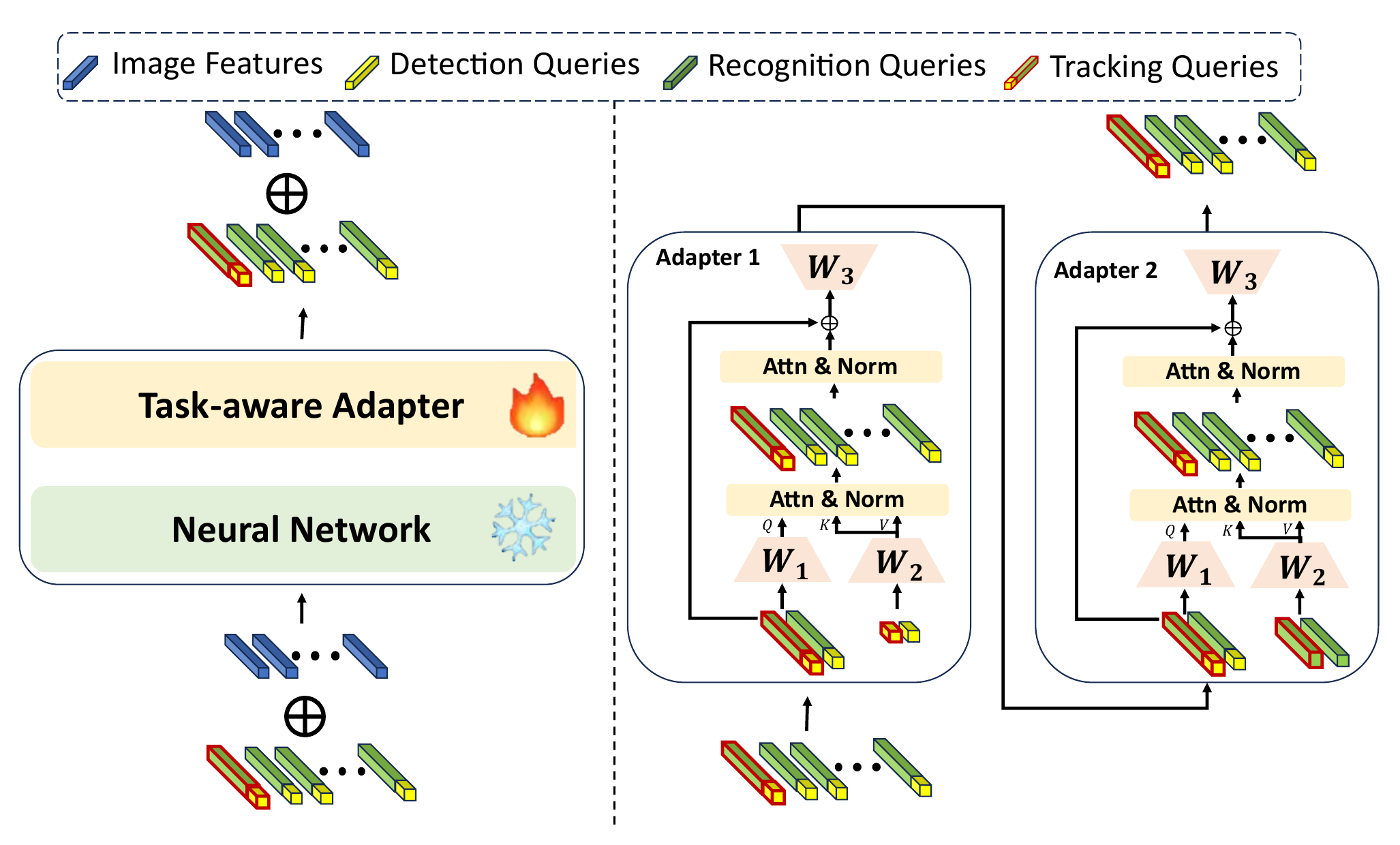}
    \caption{
        The overall structure of the task-aware adapter. The Adapter-1 is used to aggregate detection information and learn temporal information. The Adapter-2 is used to aggregate recognition information and learn temporal information.
    }
    \label{fig:adapter}
\end{figure}

Inspired by~\cite{houlsby2019parameter}, we propose a Task-aware Adapter that dynamically selects features tailored for different tasks. Combined with the PQGM, this adapter transforms the original single-task model into a multi-task model capable of handling both image and video scenarios, with minimal additional parameters. The adapter uses a cascade structure: one adapter encodes detection information, while another encodes recognition. To implement this, we freeze most parameters of a pre-trained text spotter and integrate the adapter into the neural network, such as a deformable transformer layer. During multi-task training, the adapter learns to adapt to the characteristics of each task.
The task-aware adapter is effective not only in image-level tasks but also in learning temporal information, enabling the transition of pre-trained models to video text spotting tasks. As shown in Fig.~\ref{fig:adapter}, we first use two linear layers to reduce the dimensionality of group queries to $C/4$, minimizing the parameters of the subsequent module. Task-aware queries then aggregate detection features via an attention mechanism. For image-level spotting, the attention learns relationships between text instances, while for video-level tasks, it captures temporal information. The second adapter follows a similar process, focusing on recognition. This approach efficiently learns interactive and discriminative features with minimal additional parameters, as validated in the ablation study (Sec.~\ref{subsec:ab}).

\subsection{Tracking Queries}
\label{sec:tracking_queries}
Inspired by MOTR~\cite{zeng2022motr} and ColTrack~\cite{liu2023collaborative}, we incorporate tracking queries to enable the model to trace text.
This approach enables VimTS to adapt dynamically to both images and videos. For the initial video frame, detection and recognition queries are used to locate and recognize text instances, which are then fed into the decoder to aggregate features and simultaneously decode detection and recognition results. Once aggregated, the queries encode both semantic and location information. Since text instances across consecutive frames are strongly correlated, tracking queries, based on prior frame detection and recognition results, are used for subsequent frames to track and recognize the same instances. For new text instances, detection and recognition queries are used as usual.

The overall process is shown in Fig~\ref{fig:overall}. The different queries explicitly model the discriminative and interactive features for different tasks in a single decoder and output the detection, recognition, and tracking results simultaneously. The previous end-to-end method~\cite{wu2022end} employs a relatively separate recognition head after the tacking process. Therefore, the recognition information between the different frames is not fully utilized. By benefiting from a unified framework and incorporating group queries, our method allows for the effective utilization of recognition information across various frames during the tracking process. The recognition information of previous frames can be used to help recognize the text instances in later frames, which can not be achieved by the previous end-to-end method.

\begin{table*}[!t]\small
\centering
\caption{Statistics of sourced video datasets of the VTD-368k. The Duration represents the average duration of the video. The Remaining indicates the proportion of the selected frames. Examples of the VTD-368k can be found in Tab.~\ref{fig:vtd_368k}}
\resizebox{0.85\linewidth}{!}{%
\begin{tabular}{ccccccc}
\hline
\setlength{\tabcolsep}{0.4mm}
\multirow{1}{*}{Datasets} & \multirow{1}{*}{Videos} & \multirow{1}{*}{Frames} & \multirow{1}{*}{Duration} & \multirow{1}{*}{Theme} & \multirow{1}{*}{Priority} & \multirow{1}{*}{Remaining\%}\\
\hline
\multirow{1}{*}{NExT-QA~\cite{xiao2021next}} & 2.1k & 2.1m & 50s &  kid, concerts, pets, gatherings & high & 14.1\\
\multirow{1}{*}{A2D~\cite{xu2015can}} & 2.9k & 402k & 5s & human and animal activities & high & 12.1\\
\multirow{1}{*}{ActorShift~\cite{zhang2022audio}} & 146 & 33k & 15s & animal activities & middle & 18.4\\
\multirow{1}{*}{Charades-Ego~\cite{sigurdsson2018actor}} & 2.5k & 2.0M & 50s & daily indoors activities & middle & 0.3\\
\hline
\end{tabular}}
\label{tab:dataset_information}
\end{table*}

\subsection{Optimization}

    \begin{figure*}[t!]
        \centering
        \includegraphics[width=0.85\linewidth]{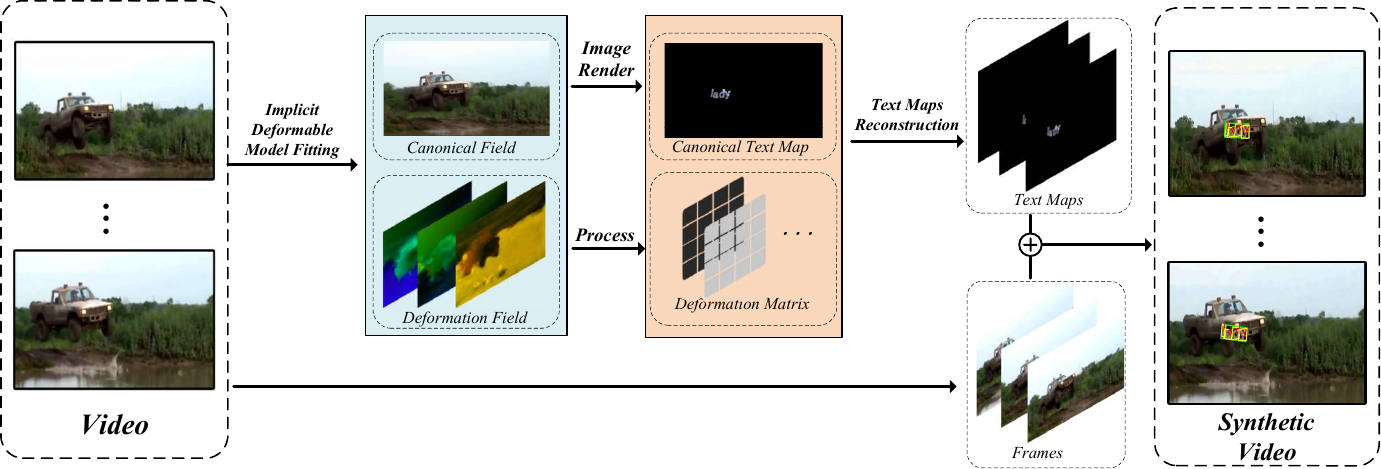}
        \caption{Overall framework of the proposed synthetic method.}
        \label{fig:synthtic_framework}
    \end{figure*}

    \begin{figure*}[t!]
        \centering
        \includegraphics[width=\linewidth]{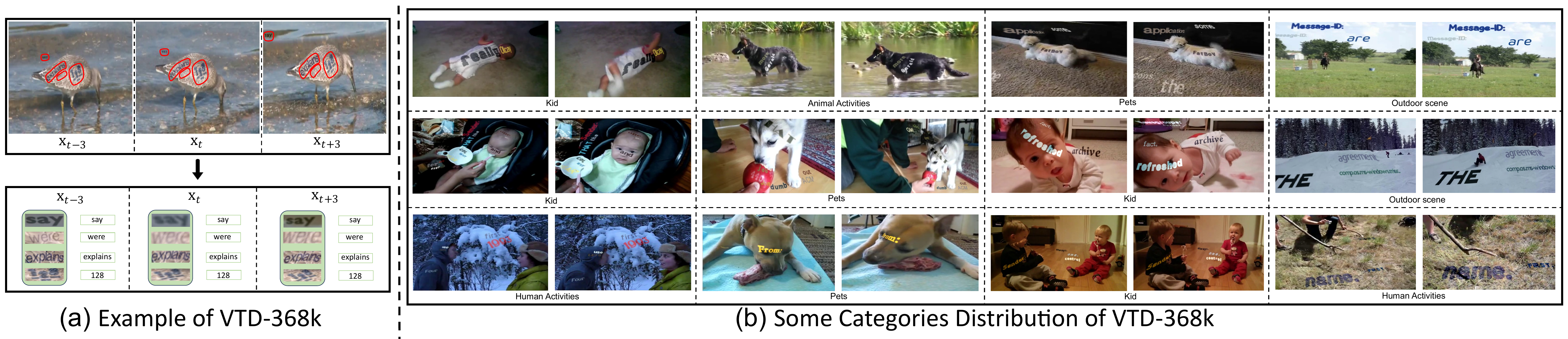}
        \caption{Distributions of VTD-368k. (a) Some annotations examples of VTD-368k. (b) Some categories distribution of VTD-368k. The synthetic dataset covers a wide scenes. Our synthetic data has a strong continuity between different frames by facilitating the achievement of realistic and stable text flow propagation.}
        \label{fig:vtd_368k}
    \end{figure*}

\textbf{Loss.} The training process of VimTS is a set prediction problem that uses a fixed number of outputs to match the ground truths. Inspired by the DETR-like methods~\cite{carion2020end,zhu2020deformable,liu2022dabdetr}, we utilize the Hungarian algorithm~\cite{kuhn1955hungarian} to perform pairwise matching and minimize the prediction-ground truth matching cost $\mathcal{C}_{match}$ as:
\begin{equation}
\hat \sigma = \underset{\sigma}{\arg \min} \sum_{i=1}^{N}\mathcal{C}_{match}(Y_i, \hat Y_{\sigma(i)})\,,
\end{equation}
where $Y_i$ is the ground truth and $\hat Y_{\sigma(i)}$ is the prediction. $N$ is the number of the predictions indexed by $\sigma(i)$. The cost function $\mathcal{C}_{match}$ is defined as:
\begin{equation}
\mathcal{C}_{match}(Y_i, \hat Y_{\sigma(i)}) = \lambda_{c}\mathcal{C}_{c}(\hat p_{\sigma(i)}(c_i)) + \mathds{1}_{\left\{c_i \neq \emptyset\right\}} \lambda_{b}\mathcal{C}_{b}(b_i, \hat b_{\sigma(i)})\,,
\end{equation}
where $c_i$ and $b_i$ are the ground truth class and bounding box, and $\hat b_{\sigma(i)}$ represents the prediction of bounding box. $\hat p_{\sigma(i)}(c_i)$ is the probability of prediction for class $c_i$. $\lambda_{c}$ and $\lambda_{b}$ are the weights for the classification and bounding box. After the Hungarian algorithm, the prediction and ground truth can be one-to-one matched. The training loss is as follows:
\begin{multline}
    \mathcal{L}(Y_i, \hat Y_{\sigma(i)}) = \alpha_{c}\mathcal{L}_{c}(\hat p_{\sigma(i)}(c_i)) + \mathds{1}_{\left\{c_i \neq \emptyset\right\}} \mathcal{L}_{b}(b_i, \hat b_{\sigma(i)}) \\
     + \mathds{1}_{\left\{c_i \neq \emptyset\right\}} \alpha_{r}\mathcal{L}_{r}(r_i, \hat r_{\sigma(i)}) + \mathds{1}_{\left\{c_i \neq \emptyset\right\}} \alpha_{p}\mathcal{L}_{p}(p_i, \hat p_{\sigma(i)})\,,
\end{multline}
where $\alpha_{c}$, $\alpha_{b}$, $\alpha_{p}$, and $\alpha_{r}$ are the loss weights for the classification, bounding box, polygon, and recognition, respectively. The classification loss $\mathcal{L}_{c}$ is the focal loss~\cite{lin2017focal}. The bounding box loss $\mathcal{L}_{b}$ consists of the 
$\ell_1$ loss and the GIoU loss~\cite{rezatofighi2019generalized}. The polygon loss uses the $\ell_1$ loss as well. The recognition loss is the standard cross entropy loss. Similar to ~\cite{zeng2022motr}, for video text spotting, the loss of newborn text instances is similar to that used in image text spotting and the text instances undergoing tracking are directly associated with the corresponding instances for loss computation.

\subsection{Synthetic Data}
Video text spotting data is highly costly. BOVText~\cite{wu2021bilingual} reports that annotating 2,021 videos requires the dedicated efforts of 30 staff members over a period of three months. Furthermore, dataset copyright is also limited to the large-scale construction of video text spotting data. Therefore, employing low-cost synthetic data is an effective method for alleviating the data requirement of video text spotting models. A solution for data synthesis is using optical flow estimation, but it poses several challenges, including distortion, labeling errors, and a bias towards static objects~\cite{10219970}. To tackle these challenges, we introduce a new approach that utilizes the CoDeF~\cite{ouyang2023codef} to facilitate the achievement of realistic and stable text flow propagation for constructing a synthetic video text dataset.

\subsubsection{Data Source and Preprocessing}
The videos used for synthesizing data must not contain text of
their own and publicly available without copyright issues. Additionally, the video should also be relatively clear, ensuring the quality of the synthesized dataset. According to this requirement, we carefully selected the video dataset used in the paper from publicly available datasets.

We exclude text-containing video datasets and manually filter single-thesis, clustered-scene, and low-resolution public video datasets. We found that the unsuitable video dataset for synthesis can be broadly categorized into four categories: 1) Text-Containing Datasets with labeling the text: These include OTB2015~\cite{wu2015object}, ActivityNet-Caption~\cite{krishna2017dense}, and AVA~\cite{gu2018ava}. The presence of unlabeled text in these videos can confuse the model during the training phase when it encounters synthesized data. 2) Single-Theme Datasets: Examples such as EPIC-KITCHENS~\cite{yang2021epic} and MPI-Cooking~\cite{rohrbach2016recognizing} lack the thematic diversity required to improve the model's generalization capabilities. 3) Cluttered-Scene Datasets: Videos from collections like Hollywood~\cite{marszalek2009actions} and Breakfast~\cite{kuehne2014language} do not offer clear areas suitable for rendering, due to their complex and cluttered scenes. 4) Low-Resolution Datasets: Such as TinyVIRAT~\cite{demir2021tinyvirat}, these datasets fail to provide the necessary visual clarity and detail for high-quality synthesis. Following this evaluation, we selected a subset of text-free, open-source, and unrestricted videos from NExT-QA~\cite{xiao2021next}, A2D~\cite{xu2015can}, ActorShift~\cite{zhang2022audio}, and Charades-Ego~\cite{sigurdsson2018actor}.

The statistics of each dataset are shown in Tab.~\ref{tab:dataset_information}. To achieve distributed processing and GPU memory reduction, we split the videos into segments that contain 368k frames for data synthesis. The synthetic data is termed VTD-368k.

In the VTD-368k dataset, the average text length is 4.14 characters, with an average of 53.45 characters appearing per frame. We define text density as the ratio of the number of text instances to the total number of frames. According to this definition, the text density of VTD-368k is 6.44. Additionally, the text styles within VTD-368k are kept consistent with those found in SynthText~\cite{gupta2016synthetic}.

\subsubsection{Synthetic Method}
Before describing the framework, we first give a brief introduction to CoDeF~\cite{ouyang2023codef}.

\textbf{CoDeF (Content Deformation Fields)}. CoDeF is an efficient way to represent a video $\textbf{V}$ comprised of
frames $\{I_1,I_2,...,I_N\}$ with a flattened canonical image  $\textbf{C}$, and a deformation field $\textbf{D}$. The process can be described as:
\begin{equation}
\{\textbf{C},\textbf{D}\} = \mathbb{F}([I_1,I_2,...,I_N]),
\label{Eq:codef_fit}
\end{equation}
where $\mathbb{F}$ denotes the fitting process of the implicit model. 
By employing specific tools such as ControlNet~\cite{zhang2023adding}, SAM~\cite{kirillov2023segment}, or R-ESRGAN~\cite{wang2021real}, to transform the canonical image $\textbf{C}$ into $\textbf{C\textquoteright}$, and concurrently integrating this transformation with the deformation field $\textbf{D}$, we can achieve video style translations, video object tracking, and video super-resolution. 
The reconstruction process for video $\textbf{V'}$ can be formulated as follow:
\begin{equation}
\textbf{V'} = \mathbb{C}(\textbf{C'},\textbf{D}),
\label{eq:codef_reconstruct}
\end{equation}
where $\mathbb{C}$ is the reconstructing process of the implicit model and $\textbf{V'}$ is the reconstructed video.

\textbf{Our method}. We propose a new synthetic method. Specifically, as shown in Fig.~\ref{fig:synthtic_framework}, we first use SAM-Track~\cite{cheng2023segment} and RAFT~\cite{teed2020raft} to obtain the optical flow maps and segmentation maps of all frames, respectively. Then, the CoDeF is used to effectively reconstruct both rigid and non-rigid objects within videos, while meticulously restoring the subtle intricacies of motion details. 
Processed by the CoDeF, an input video is represented as a canonical image $\textbf{C}$ and deformation field $\textbf{D}$. Where, $\textbf{C}$ serves as the substrate for text placement via Synthtext, while $\textbf{D}$ encapsulates the transformation from observation to canonical representation for each frame. Subsequently, the embedding text map $\bm{T_c}$ can be generated as follows:

\begin{equation}
\{\bm{C},\bm{T_c}\} = \underset{\text{Render}}{\mathbb{M}}(\bm{C})|_{\{S_c,D_c\}},
\end{equation}
where ${S_c}$ and ${D_c}$ denote segmentation map and depth map of canonical image $\textbf{C}$. Then, $\bm{T_c}$ is used as the input of Eq.~\ref{eq:codef_reconstruct} and combined with $\bm{D}$ to reconstruct video text maps $\bm{T_t}$, using the following equation:
\begin{equation}
\bm{T_t} = \mathbb{C}(\bm{T_c},\textbf{D}).
\end{equation}

The trained implicit deformable model generates text geometry, but it does not preserve its collinearity and stroke sequence. To address this issue, the projective transformation is introduced after the reconstruction of the implicit deformable model. Specifically, we collect a set of point pairs $(\bm{p}_c,\bm{p}_t)$. Where $\bm{p}_c$ are the points in the canonical image that form a text geometry, and $\bm{p_t}$ are their corresponding points in the reconstructed frames. We then estimate the projective matrix $H_{c,t}$ by RANSAC~\cite{fischler1981random} to fit those point pairs robustly. Finally, we apply $H_{c,t}$ to transform each text geometry in the text maps.

\begin{table*}[t]
\setlength\tabcolsep{1pt}
\centering
\caption{Real-to-real adaptation on scene text detection methods.
Detection metric `F-measure' ($\%$) is reported. $TT_W$ represents the word-level TotalText. $TT_L$ represents the line-level TotalText. $CTW_W$ represents the word-level CTW1500. $CTW_L$ represents the line-level CTW1500. Bold indicates SOTA. The same conventions apply to the table below. 
}
\setlength{\tabcolsep}{0.7pt}
\resizebox{0.9\linewidth}{!}{
\newcommand{\tabincell}[2]{\begin{tabular}{@{}#1@{}}#2\end{tabular}}
\begin{tabular}{lccccccccc}
    \toprule
	\multirow{1}[0]{*}{Method} & 
    \multicolumn{1}{c}{Venue} & 
	\multicolumn{1}{c}{$TT_W$ $\rightarrow$ $IC15$} &
	\multicolumn{1}{c}{$TT_W$ $\rightarrow$ $CTW_L$} &
    \multicolumn{1}{c}{$IC15$ $\rightarrow$ $CTW_L$} &
	\multicolumn{1}{c}{$CTW_L$ $\rightarrow$ $IC15$} & 	 
    \multicolumn{1}{c}{$CTW_L$ $\rightarrow$ $TT_L$} & 
    \multicolumn{1}{c}{$TT_W$ $\rightarrow$ $CTW_W$} & \multicolumn{1}{c}{Average} \\
     \midrule
    ABCNet v2~\cite{liu2021abcnetv2} & TPAMI'2021 & 80.2  & 46.4 & 46.5 & 45.0 & 70.1 & 88.5 & 62.8\\
	TESTR~\cite{zhang2022text} & CVPR'2022 & 79.8  & 48.1 & 46.6 & 46.4 &72.8 &87.5 & 63.5\\
    TPSNet~\cite{wang2022tpsnet} & ACMMM'2022 & 78.4 &  46.0 & 42.7 & 42.6 & 71.7 & 88.3 & 61.6\\
    ABINet++~\cite{fang2022abinet++} & TPAMI'2022 & 79.8  &  47.7 & 49.0 & 41.8 & 71.6 & 88.1 & 63.0\\
    DeepSolo~\cite{ye2022dptext} & CVPR'2023 & 79.7  &  46.9 & 47.2 & 50.0 & 72.5 & 89.6 & 64.3\\
    ESTextSpotter~\cite{huang2023estextspotter}  & ICCV'2023 & 83.4  & 48.0 & 46.3 & 55.6 & 76.8 & 90.4 & 66.8\\
    FastTCM~\cite{yu2023turningts} & TPAMI'2024 & 81.1 &  47.8 & 46.9 & 53.7 & 75.9 & 87.6 & 65.5\\
    \hline
    VimTS  & - & \textbf{85.5}  & \textbf{48.1} & \textbf{47.9} & \textbf{56.2} & \textbf{77.4} & \textbf{91.5} & \textbf{67.8}\\
    \bottomrule
\end{tabular}
}
\label{tab:cp_sota_det}
\end{table*}

\begin{table*}[t]
\setlength\tabcolsep{1pt}
\centering
\caption{Real-to-real adaptation on scene text spotting methods.
End-to-End spotting metric `None' ($\%$) is reported. 
}
\setlength{\tabcolsep}{0.7pt}
\resizebox{0.9\linewidth}{!}{
\newcommand{\tabincell}[2]{\begin{tabular}{@{}#1@{}}#2\end{tabular}}
\begin{tabular}{lccccccccc}
    \toprule
	\multirow{1}[0]{*}{Method} & 
    \multicolumn{1}{c}{Venue} & 
	\multicolumn{1}{c}{$TT_W$ $\rightarrow$ $IC15$} &
	\multicolumn{1}{c}{$TT_W$ $\rightarrow$ $CTW_L$} &
    \multicolumn{1}{c}{$IC15$ $\rightarrow$ $CTW_L$} &
	\multicolumn{1}{c}{$CTW_L$ $\rightarrow$ $IC15$} & 	 
    \multicolumn{1}{c}{$CTW_L$ $\rightarrow$ $TT_L$} & 
    \multicolumn{1}{c}{$TT_W$ $\rightarrow$ $CTW_W$} & 
    \multicolumn{1}{c}{Average}\\
     \midrule
    ABCNet v2~\cite{liu2021abcnetv2} & TPAMI'2021 & 54.3  & 27.4 & 21.7 & 24.5 & 46.5 & 75.9 & 41.7\\
	TESTR~\cite{zhang2022text} & CVPR'2022 & 55.9  & 29.0 & 26.6 & 23.8 &44.4 &76.9 & 42.8\\
    TPSNet~\cite{wang2022tpsnet} & ACMMM'2022 & 60.5  & 27.5 & 20.1 & 23.2 &41.8 &77.8 & 41.8\\
    ABINet++~\cite{fang2022abinet++} & TPAMI'2022  & 62.0  &  29.7 & \textbf{30.4} & 20.3 & 43.5 & 78.8 & 44.1 \\
    DeepSolo~\cite{ye2022dptext} & CVPR'2023 & 63.6  &  29.6 & 26.7 & 25.0 & 50.5 & 81.7 & 46.2\\
    SPTS v2~\cite{liu2023spts} & TPAMI'2023 & 58.7  &  28.5 & 28.5 & 33.2 & 52.7 & 79.5 & 46.9\\
    ESTextSpotter~\cite{huang2023estextspotter}  & ICCV'2023 & 65.0  & 29.5 & 27.3 & 31.6 & 50.5 & 80.9 & 47.5\\
    FastTCM~\cite{yu2023turningts} & TPAMI'2024 & 56.7  &  28.6 & 26.1 & 25.7 & 45.3 & 77.2 & 43.2\\
    \hline
    VimTS  & - & \textbf{68.0}  & \textbf{30.4} & 29.6 & \textbf{35.9} & \textbf{54.5} & \textbf{82.2} & \textbf{50.1}\\
    \bottomrule
\end{tabular}
}
\label{tab:cp_sota_e2e}
\end{table*}

By utilizing the CoDeF, our method facilitates the achievement of realistic and stable text flow propagation, significantly reducing the occurrence of distortions.  Furthermore, it imposes minimal constraints during the text flow propagation process, effectively mitigating the risk of ignoring dynamic objects. 

Despite the numerous benefits that CoDeF offers, it does come with certain limitations. Specifically, the CoDeF-based synthesis method relies on an image rendering algorithm to process canonical image $\textbf{C}$ generated by an implicit deformable model. However, it is difficult to use the deformation field $\textbf{D}$ and the renderable canonical image $\textbf{C}$ to represent those videos with dynamic camera movement. Therefore, inspired by~\cite{10219970}, we propose a new method to handle those videos via optical flow estimation. The original method in~\cite{10219970} produces distortion in the text flow propagation process due to the long-distance accumulation of optical flow, which causes abnormal sampling points and leads to erroneous computation of the homography matrix. To achieve more stable synthesis results, we apply Eq.~\ref{eq:forward} and Eq.~\ref{eq:backward} to implement the frame-by-frame forward and backward text flow propagation process of the text flow:
\begin{equation}
p_{k+1} = \mathit{F}_{k \Leftrightarrow k+1}(p_k)+p_k, p_k \in T_k , k \in [t,N-1),
\label{eq:forward}
\end{equation}
\begin{equation}
p_{t-k-1} = \mathit{F}_{t-k \Leftrightarrow t-k-1}(p_k)+p_{t-k}, p_{t-k} \in T_{t-k} , k \in [0,t),
\label{eq:backward}
\end{equation}
where $t$ is the id of the rendered frame. Similarly, we also perform character-level text flow propagation and assign correct labels to the text maps in each frame. Some examples of VTD-368k are visualized in Fig.~\ref{fig:vtd_368k}.

\section{Experiments on Image Text Spotting}
For image-level cross-domain text spotting, following~\cite{yu2023turningts}, we conduct experiments on common cross-domain settings to evaluate VimTS, including arbitrarily shaped datasets TotalText~\cite{ch2019total} and CTW1500~\cite{liu2019curved}, and multi-oriented dataset ICDAR2015~\cite{karatzas2015icdar}.

\subsection{Implementation Details}

We pre-train the model on a combination of Curved SynthText~\cite{liu2020abcnet}, ICDAR-MLT~\cite{nayef2017icdar2017}, and the corresponding datasets with $240$k iterations. The base learning rate is $1 \times 10^{-4}$ and reduced to $1 \times 10^{-5}$ at the $180$k-th iteration and $1 \times 10^{-6}$ at $210$k-th iteration. Then, the model is fine-tuned on the corresponding real datasets. We use $N=100$ as the maximum number of predictions. The max length of recognition queries is $25$. The weight for the classification loss $\alpha_{c}$ is $2.0$. The weight of the $\ell_1$ loss is $5.0$ and the GIoU loss is $2.0$.
The polygon loss weight $\alpha_{p}$ and the recognition loss weight $\alpha_{r}$ are both set to $1.0$. The focal loss parameters $\alpha$ and $\gamma$ are $0.25$ and $2.0$, respectively. The number of both encoder and decoder layers is 6. We obtain the TotalText-line and CTW1500-word by manually labeling, which has been made open-sourced.

The data augmentation strategies used 
are also kept the same as previous works~\cite{zhang2022text,liu2020abcnet,liu2021abcnetv2} as follows: 1) random resizing with the shorter size chosen from $640$ to $896$ pixels (with an interval of $32$), and the longest size is constrained within $1600$ pixels; 2) random cropping, which ensures that text is not being cut; 3) random rotation, which rotates the images with an angle in the range of $[-45 ^ {\circ}$, $45 ^ {\circ}]$. For testing, we resize the shorter size of the image to $1000$ pixels while keeping the longest size of the image within $1824$ pixels. All experiments are conducted on 8 NVIDIA RTX3090 GPUs.

For the `Zero-shot' setting, training requires approximately 45 hours on 8 RTX 3090 GPUs. The `with VTD-368k' setting takes around 80 hours, and `with video data' requires about 90 hours on the same setup.

\begin{table*}[t]
\setlength\tabcolsep{1pt}
\centering
\caption{Ablation study of unified multi-granularity fine-tuning methods. The results are tested on Real-to-real adaptation on scene text spotting methods. End-to-end spotting metric `None' ($\%$) is reported. The Full-Tuning represents tuning all the parameters of the model. PQGM represents the Prompt Queries Generation Module.
}
\setlength{\tabcolsep}{0.8pt}
\resizebox{0.85\linewidth}{!}{
\newcommand{\tabincell}[2]{\begin{tabular}{@{}#1@{}}#2\end{tabular}}
\begin{tabular}{lcccccccccc}
    \toprule
	\multirow{1}[0]{*}{Method} & \multicolumn{1}{c}{Tuned Params} & Full-Tuning & PQGM & Adapter & TA Adapter & 
	\multicolumn{1}{c}{$TT_W$ $\rightarrow$ $IC15$} &
    \multicolumn{1}{c}{$CTW_L$ $\rightarrow$ $TT_L$} & 
    \multicolumn{1}{c}{$TT_W$ $\rightarrow$ $CTW_W$} & Training Time\\
     \midrule
    Baseline & - & - & - & - & -   & 65.0 & 50.5 & 80.9 & -\\
    Baseline+  & 56.6M (100\%) & \checkmark  & - & - & -   & 62.5 (-2.5) & 36.6 (-13.9) & 71.1 (-9.8) & 210 min \\
    Baseline+  & 56.9M (100\%) & \checkmark & \checkmark & - & -  & 67.6 (+2.6)  & 54.0 (+3.5) & 81.5 (+0.6) & 210 min\\
    Baseline+ & 0.7M (1.2\%) & - & \checkmark & \checkmark & -  & 67.2 (+2.2) & 53.9 (+3.4) & 81.3 (+0.4)  & 140 min \\
    Baseline+ & 1.7M (3.0\%) & -  & \checkmark & - & \checkmark  & 68.0 (+3.0) & 54.5 (+4.0) & 82.2 (+1.3) & 140 min \\
    \bottomrule
\end{tabular}
 }
 \label{tab:cross_domain_ab}
\end{table*}

\subsection{Comparison with State-of-the-arts Methods}

\textbf{Cross-Domain Text Spotting in Multi-Oriented Text.} We evaluated the robustness of our method on the multi-oriented text by conducting experiments on the ICDAR2015 dataset. Following the approach~\cite{yu2023turningts}, we considered two main settings: 1) training the model on the word-level dataset TotalText and testing it on ICDAR2015, and 2) training the model on the line-level dataset CTW1500 and testing it on ICDAR2015. The detection results are presented in Tab.~\ref{tab:cp_sota_det}. Our method achieves the highest H-mean score of $85.5\%$ on the ICDAR2015 dataset using TotalText's model, outperforming ESTextSpotter by $2.1\%$. The end-to-end recognition results on ICDAR2015 are shown in Tab.~\ref{tab:cp_sota_e2e}. Using the CTW1500's model to evaluate ICDAR2015, our method outperforms DeepSolo by $10.9\%$, demonstrating the effectiveness of the proposed method. Additionally, our method surpasses the previous cross-domain method FastTCM by $4.4\%$ using TotalText's model to evaluate ICDAR2015.

\textbf{Cross-Domain Text Spotting in Arbitrarily-Shaped Text.} Following the previous method~\cite{yu2023turningts}, we test our method on two arbitrarily-shaped text benchmarks (TotalText and CTW1500) to verify the generalization ability of our method for arbitrarily-shaped cross-domain scene text spotting. For the text detection task, as shown in Tab.~\ref{tab:cp_sota_det}. VimTS outperforms the previous methods with $91.5\%$ in terms of the H-mean metric on line-level TotalText dataset, $1.9\%$ higher than the DeepSolo. The result of cross-domain text spotting is presented in Tab.~\ref{tab:cp_sota_e2e}. Notably, even without line-level annotations from TotalText for training, our method demonstrates promising performance in line-level Totaltext. Furthermore, our method outperforms the previous cross-domain method, FastTCM, by an average of $6.9\%$ in six cross-domain benchmarks. For text detection, our method outperforms the state-of-the-art approach by an average of $1.0\%$ across six cross-domain benchmarks. Similarly, in text spotting, our method demonstrates superiority over the state-of-the-art by an average of $2.6\%$ across the same benchmarks. These results demonstrate the strong generalization ability of our method.

\subsection{Ablation Study}
\label{subsec:ab}
We conduct ablation studies in cross-domain text spotting using TotalText-line, ICDAR2015, and CTW1500-word to evaluate the effectiveness of the proposed method. We directly train the model using both TotalText and CTW1500.

\textbf{Combining Datasets with Distinct Annotation Formats.}
We first attempt to directly unify the word-level and line-level text spotting into a single framework without the PQGM. However, as shown in the second line of Tab.~\ref{tab:cross_domain_ab}, the result demonstrates that combining the hierarchical text spotting with annotations in different formats at the same time will disturb the training of the model, reducing the performance of the model. 

\textbf{Performing Task Interaction to Achieve Task Synergy.}
To effectively unify the word-level and line-level text spotting, we propose a Prompt Queries Generation Module (PQGM) to guide the model in completing the corresponding tasks. The result is presented in the third line of Tab.~\ref{tab:cross_domain_ab}. With the PQGM, $5.1\%$, $17,4\%$, and $10.4\%$ improvement are achieved in $TT_W$ $\rightarrow$ $IC15$, $CTW_L$ $\rightarrow$ $TT_L$, and $TT_W$ $\rightarrow$ $CTW_W$, respectively. These results demonstrate the effectiveness of the proposed PQGM in unifying the hierarchical text spotting. PQGM facilitates model adaptation to both word-level and line-level text spotting, enhancing the synergy between hierarchical text spotting rather than compromising performance.

Additionally, it’s noteworthy that fine-tuning the model with the respective real datasets typically takes about 3.5 hours. However, using PQGM with the task-aware adapter can further reduce this fine-tuning time to approximately 140 minutes, as shown in Tab.~\ref{tab:cross_domain_ab}.

\textbf{Effectiveness of the Task-Aware Adapter.}
To verify the effectiveness of the task-aware Adapter, we conducted experiments to compare it with the original Adapter and the full-finetuning method. As indicated in the fourth line of Tab.~\ref{tab:cross_domain_ab}, using the original Adapter for fine-tuning the pretrained model resulted in performance degradation compared to full fine-tuning. In contrast, our method outperforms full fine-tuning with minimal trainable parameters, demonstrating its effectiveness. We can observe that compared to the full-finetuning method, using the task-aware adapter improves $TT_W$ $\rightarrow$ $IC15$, $CTW_L$ $\rightarrow$ $TT_L$, and $TT_W$ $\rightarrow$ $CTW_W$ by $0.4\%$, $0.5\%$, and $0.7\%$, respectively. Furthermore, our method achieves these enhancements while only utilizing $3.0\%$ of the parameters required for fine-tuning the entire model.

\textbf{Effectiveness of the order of detection and recognition queries.}
To further validate the influence of the order, we conduct an experiment as shown in Tab.~\ref{tab:the_order}. The results show that detection before recognition is slightly better than recognition first setting. 

\begin{table}[t]
\setlength\tabcolsep{1pt}
\centering
\caption{Ablation study of the order of aggregating information in task-aware adapter. 
}
\setlength{\tabcolsep}{0.8pt}
\newcommand{\tabincell}[2]{\begin{tabular}{@{}#1@{}}#2\end{tabular}}
\begin{tabular}{lccccc}
    \toprule
	& \multirow{1}[0]{*}{Method}  & Det. First & 
	\multicolumn{1}{c}{$TT_W$-$IC15$} &
    \multicolumn{1}{c}{$CTW_L$-$TT_L$} & 
    \multicolumn{1}{c}{$TT_W$-$CTW_W$} \\
     \midrule
   & Method  & -   & 67.5 & 54.1 & 81.9 \\
   & Method  & \checkmark  & 68.0 & 54.5 & 82.2 \\
    \bottomrule
\end{tabular}
\label{tab:the_order}
\end{table}

\begin{table*}[h]
\centering
\caption{Ablation study of using different training data. ‘M-Tracked’ and ‘M-Lost’ denote ‘Mostly Tracked’ and ‘Mostly Lost’, respectively.}
\footnotesize 
\begin{tabular}{l|ccccc}
\hline
\multirow{2}{*}{Data} & \multicolumn{5}{c}{End to End Video Text Spotting/\%} \\
\cline{2-6} 
& ${
ID_{F1}}$$\uparrow$  & ${
MOTA}$$\uparrow$ & ${
MOTP}$$\uparrow$ & ${
M\mbox{-}Tracked}$$\uparrow$ & ${
M\mbox{-}Lost}$$\downarrow$ \\
\hline
VTD-368k & 53.7 & 30.9 & 74.6 & 29.1 & 47.2 \\
image-level data & 73.2 & 61.0 & 77.1 & 52.2 & 25.7 \\
image-level data + VTD-368k  &  74.1 & 61.8 & 77.3 & 52.7 & 25.6 \\
image-level data + ICDAR2015 &  74.5 & 62.0 & 77.5 & 52.6 & 25.2 \\
image-level data + ICDAR2015 +  VTD-368k &  \textbf{75.0} & \textbf{62.3} & \textbf{77.8} & \textbf{52.9} & \textbf{25.3} \\
\hline
\end{tabular}
\label{tab:syn_real}
\end{table*}

\begin{table}[!t]\small
\caption{Video text detection results in ICDAR2013 video dataset. The video data represent the training dataset of ICDAR2015 video.}
\centering
\footnotesize
\begin{tabular}{l|ccc}
\hline
\multirow{2}{*}{Method} & \multicolumn{3}{c}{Video Text Detection/\%} \\
\cline{2-4} 
& Precision  & Recall & F-measure \\
\hline
Epshtein \textit{et al.}~\cite{epshtein2010detecting} & 39.8 & 32.5 & 35.9 \\
Zhao \textit{et al.}~\cite{zhao2010text} & 47.0 & 46.3 & 46.7 \\
Yin \textit{et al.}~\cite{yin2013robust} & 48.6 & 54.7 & 51.6 \\
Khare \textit{et al.}~\cite{khare2017arbitrarily} & 57.9 & 55.9 & 51.7 \\
Wang \textit{et al.}~\cite{wang2018scene} & 58.3 & 51.7 & 54.5 \\
Shivakumara \textit{et al.}~\cite{shivakumara2017fractals} & 61.0 & 57.0 & 59.0 \\
Wu \textit{et al.}~\cite{wu2015new} & 63.0 & 68.0 &  65.0 \\
Yu \textit{et al.}~\cite{yu2021end} & 82.4 & 56.4 & 66.9 \\
Wei \textit{et al.}~\cite{feng2021semantic} & 75.5 & 64.1 & 69.3 \\
Free~\cite{cheng2020free} & 79.7 & 68.4 & 73.6 \\
TransDETR~\cite{wu2022end} & 80.6 & 70.2 & 75.0 \\
VLSpotter~\cite{Zu2023vlspotter} & 82.3 & 71.8 & 76.0 \\
\hline
VimTS (Zero-shot) & 81.2 & 72.3 & 76.5 \\
VimTS (with VTD-368k) & 81.5 & 73.5 & 77.3 \\
VimTS (with video data)  & \textbf{83.2} & \textbf{74.1} & \textbf{78.4} \\
\hline
\end{tabular}
\label{tab:13det}
\end{table}

\begin{table}[!t]\small
\centering
\caption{Video text detection results in DSText v2. The video data represents the training dataset of DSText v2.}
\footnotesize
\begin{tabular}{l|ccc}
\hline
\multirow{2}{*}{Method} & \multicolumn{3}{c}{Video Text Detection/\%} \\
\cline{2-4} 
& Precision  & Recall & F-measure \\
\hline
EAST ~\cite{zhou2017east} & 72.1 & 37.8 & 49.6 \\
PSENet ~\cite{wang2019arbitrary} & 76.1 & 38.5 & 51.1 \\
DB ~\cite{liao2020real} & 76.5 &40.3 &52.8 \\
DB++ ~\cite{liao2022real} & 77.8 &41.6 &54.2 \\
TransVTSpotter~\cite{wu2021bilingual} & 77.2 & 45.3 & 57.1  \\
TransDETR~\cite{wu2022end} & 78.5 & 52.5 & 62.9  \\
\hline
VimTS (Zero-shot) & 78.7 & 53.5 & 63.7 \\
VimTS (with VTD-368k) & 79.1 & 54.4 & 64.2 \\
VimTS (with Video data)  & \textbf{79.3} & \textbf{55.2} & \textbf{65.0}  \\
\hline
\end{tabular}
\label{tab:dsdet}
\end{table}

\begin{table}[!t]\small
\centering
\caption{Zero-shot video text spotting on ICDAR2013 video. All methods use the same training set and evaluate on each frame of the video.}
\footnotesize
\begin{tabular}{l|ccc}
\hline
\multirow{2}{*}{Method} & \multicolumn{3}{c}{Video Text Spotting/\%} \\
\cline{2-4} 
& Precision  & Recall & F-measure \\
\hline
ABCNet ~\cite{liu2020abcnet} & 49.2 & 26.5 & 34.4 \\
ABCNet v2 ~\cite{liu2021abcnetv2} & 57.8 & 31.8 & 41.0 \\
ABINet++ ~\cite{fang2022abinet++} & 55.4 & 30.2 & 39.1 \\
TESTR ~\cite{zhang2022text} & 52.4 & 36.7 & 43.2 \\
DeepSolo ~\cite{ye2023deepsolo} & 56.8 & 37.8 & 45.4 \\
FastTCM ~\cite{yu2023turningts} & 53.6 & 34.3 & 41.8 \\
\hline
VimTS & \textbf{77.3} & \textbf{41.0} & \textbf{53.6} \\
\hline
\end{tabular}
\label{tab:vtic13}
\end{table}

\section{Experiments on Video Text Spotting}

\begin{table*}[]
\centering
\caption{Video Text Spotting results. The video data represent the corresponding training data of the benchmarks.}
\footnotesize 
\begin{tabular}{l | c | c | ccccc }
\hline
\multirow{2}{*}{Method} & \multirow{2}{*}{Venue}  &\multirow{2}{*}{Backbone} & \multicolumn{5}{c}{End to End Video Text Spotting/\%} \\
\cline{4-8} 
& && ${
ID_{F1}}$$\uparrow$  & ${
MOTA}$$\uparrow$ & ${
MOTP}$$\uparrow$ & ${
M\mbox{-}Tracked}$$\uparrow$ & ${
M\mbox{-}Lost}$$\downarrow$ \\
\hline
& \multicolumn{6}{c}{ICDAR2015} \\ \hline
\textit{Separate Framework}& & & & & \\
USTB\_TexVideo(II\mbox{-}2)~\cite{karatzas2015icdar} & ICDAR'2015 & & 21.3 & 13.2 &  66.6 & 6.6& 67.7  \\
USTB\_TexVideo~\cite{karatzas2015icdar} & ICDAR'2015 & & 28.2 & 15.6 & 68.5 & 9.5& 60.7\\
StradVision\mbox{-}1~\cite{karatzas2015icdar} & ICDAR'2015 & & 32.0 & 9.0 &  70.2 & 8.9& 59.5 \\
Free~\cite{cheng2020free} & TIP'2020 & ResNet50 & 61.9 & 53.0 &  74.9 & \textbf{45.5} & 35.9 \\
 TransVTSpotter\cite{wu2021bilingual} & NuerIPS'2021 & ResNet50 & 61.5 & 53.2 &  74.9 & - & - \\
\hline
\textit{End-to-End framework}& & & & & & \\
TransDETR~\cite{wu2022end} & IJCV'2024 & ResNet50 & 72.8 & 60.9 & 74.6 & 33.6 & 20.8  \\
CoText~\cite{wu2022end} & - & ResNet18 & 72.0&  59.0 & 74.5 & 48.6 & 26.4  \\
VLSpotter~\cite{Zu2023vlspotter} & IJCAI'2023 & ResNet50 & 72.8 & 60.1 & 76.4 & 50.0 & 26.1  \\
VimTS (Zero-shot) & - & ResNet50 &  73.2 & 61.0 & 77.1 & 52.2 & 25.7 \\
VimTS (with VTD-368k) & - & ResNet50 &  74.1 & 61.8 & 77.3 & 52.7 & 25.6 \\
VimTS (with Video data) & - & ResNet50 &  \textbf{75.0} & \textbf{62.3} & \textbf{77.8} & \textbf{52.9} & \textbf{25.3} \\
\hline
& \multicolumn{6}{c}{DSText v2} \\ \hline
\textit{Separate Framework} &  & & & & & \\
EAST+VMFT+CRNN~\cite{wu2024dstext} & PR'2024 & ResNet50 & 9.3 & -31.2 & 77.3 & 553 & 11497  \\
PSENet+VMFT+CRNN~\cite{wu2024dstext} & PR'2024 & ResNet50 & 10.2 & -31.0 & 77.3 & 536 & 11454  \\
\hline
\textit{End-to-End framework}& & & & & & \\
TransDETR~\cite{wu2022end} & IJCV'2024 & ResNet50 & 12.0 & -11.9 & 78.8 & 446 & 11818  \\
VimTS (Zero-shot)  & - & ResNet50 &  12.5 & -1.0 & 82.1 & 744 & 11362  \\
VimTS (with VTD-368k) & - & ResNet50 &  13.2 &-0.5 & 82.5 & 783 & 11031  \\
VimTS (with video data) & - & ResNet50 &  \textbf{20.0} & \textbf{3.2} & \textbf{83.5} & \textbf{812} & \textbf{10024}  \\
\hline
\end{tabular}
\label{tab:video_spotting}
\end{table*}

\subsection{Implementation Details}
Following previous methods~\cite{wu2021bilingual,wu2022end}, in our experiments, we develop a two-stage training manner. We first train our model using image-level datasets, which include ICDAR2013~\cite{karatzas2013icdar}, ICDAR2015~\cite{karatzas2015icdar}, TotalText~\cite{ch2019total}, and MLT2017~\cite{nayef2017icdar2017}. Consistent with prior works~\cite{wu2021bilingual, wu2022end}, we incorporate random shift~\cite{zhou2020tracking} to generate video clips with pseudo tracks in static images. For using the video data, we utilize the proposed synthetic video data. We jointly tune the PQGM and task-aware adapter on both image-level and video-level data building upon the pre-trained model on image text spotting. We utilize the AdamW~\cite{loshchilov2017decoupled} with a learning rate of $2 \times 10^{-4}$ as the optimizer. Following previous work~\cite{wu2022end}, we implement sequential data augmentation strategies, encompassing the following steps: 1) Random resizing of the image, with the shorter dimension chosen from a range of 608 to 992 pixels (in intervals of 32), while ensuring the longest dimension does not exceed 1536 pixels; 2) Random cropping, designed to preserve the entirety of the text without any truncation.

\subsection{Zero-shot Transfer from Image to Video}
To evaluate our method's ability in video text spotting, we conduct a comparison with a previous image-level text spotting method using the ICDAR2013 video dataset~\cite{karatzas2013icdar} in a zero-shot manner. All methods are trained only using the image-level data and directly evaluated on each frame of the ICDAR2013 video in a zero-shot manner. The results are shown in Tab.~\ref{tab:vtic13}. VimTS achieves $53.6\%$ in the F-measure metrics, outperforming the previous image-level text spotting method by $8.2\%$ at least. These results demonstrate the superior generalization ability of our proposed method in the domain of video text spotting. 

\begin{figure*}[!t]
    \centering
    \begin{minipage}[c]{0.95\linewidth}
    \centering
    {\includegraphics[width=15.7cm,height=3.cm]{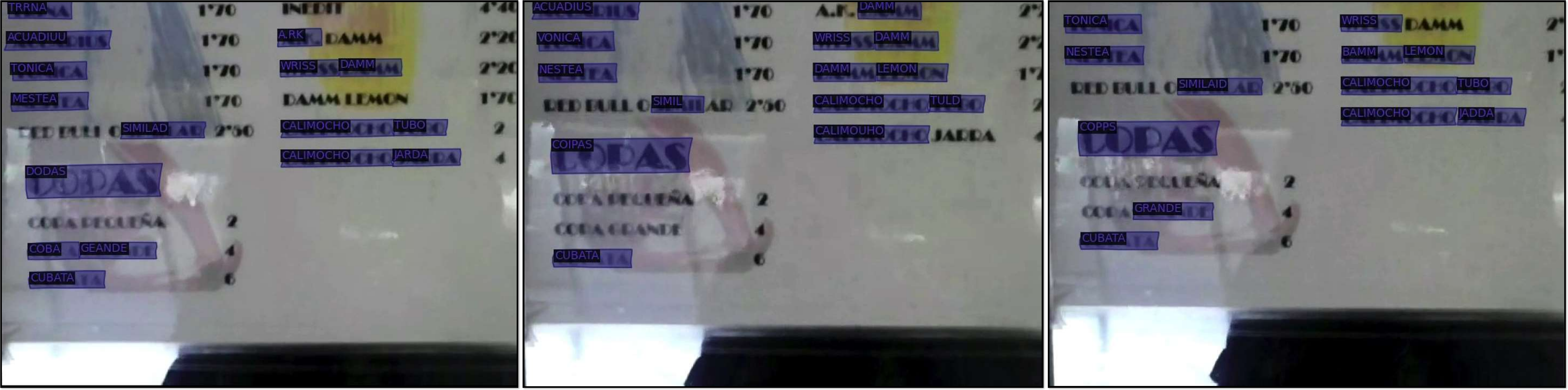}}\label{fig:vis_testr}
    \centerline{\small{(a) TESTR~\cite{zhang2022text} (zero-shot). It is trained on the image-level data.}}
    \end{minipage}
    \begin{minipage}[c]{0.95\linewidth}
    \centering
    {\includegraphics[width=15.7cm,height=3.cm]{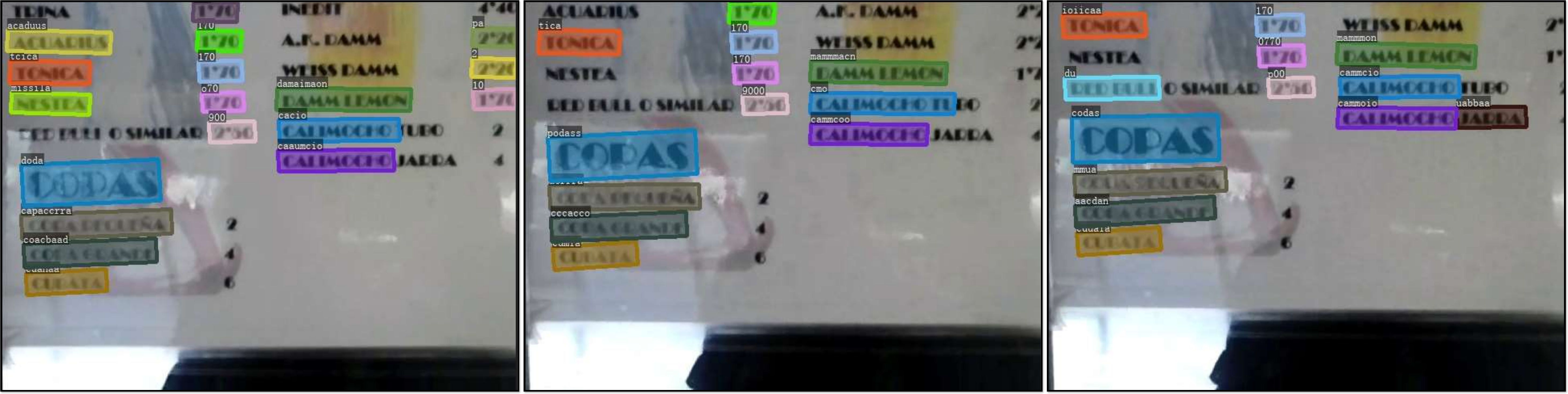}}\label{fig:vis_trans}
    \centerline{\small{(b) TransDETR~\cite{wu2022end}. It is trained on the video-level data.}}
    \end{minipage}
    \begin{minipage}[c]{0.95\linewidth}
    \centering
    {\includegraphics[width=15.7cm,height=3.cm]{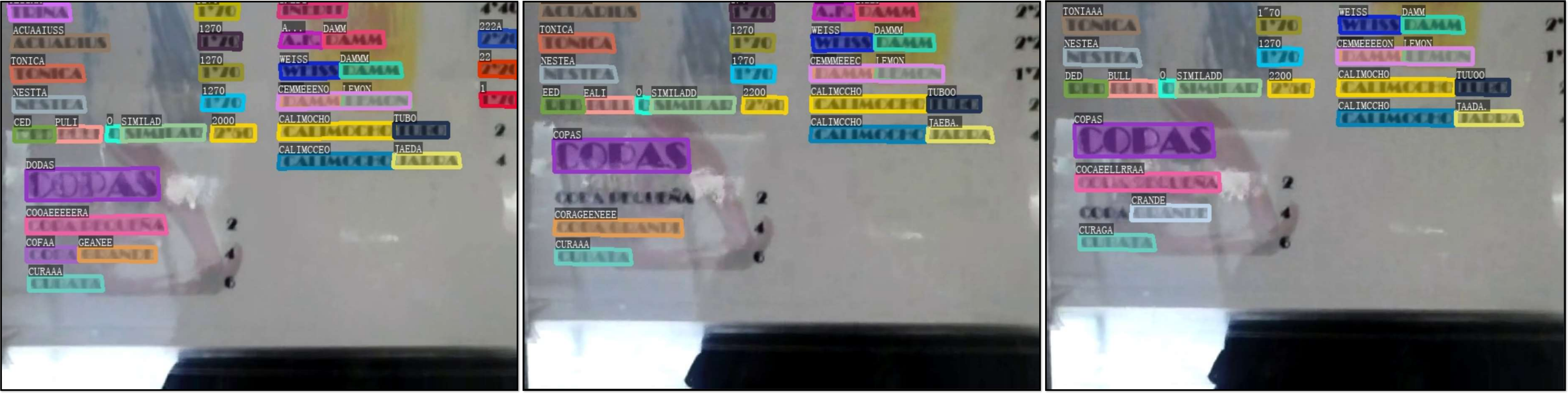}}\label{fig:video_ours}
    \centerline{\small{(c) Ours (zero-shot). It is trained on the image-level data.}}
    \end{minipage}
    \caption{Visualization results of VimTS (ours), TESTR, and TransDETR on ICDAR2015(video). The same color between different frames represents the same tracking ID. Best view on screen.}
    \label{fig:vis_results}
\end{figure*}

\subsection{Comparison with State-of-the-arts Methods}
We compare our method with state-of-the-art methods in terms of three tasks, containing video text detection, video
text tracking, and video text spotting. The evaluation metrics of video text detection, spotting, and tracking following the previous method~\cite{wu2022end}, including Precision, Recall, F-Measure, IDF1, MOTA, MOTP, M-Tracked, and M-Lost. 
We evaluate our method in three manners: 1) Training with only image-level datasets and directly evaluating the video datasets in a zero-shot manner. 2) Training with image-level data and VTD-368k, and evaluating the video datasets. 3) Training with image-level data, VTD-368k, and video data, and evaluating the video datasets.

\textbf{Video text detection.} 
For video text detection, similar to previous works~\cite{wu2022end,Zu2023vlspotter}, we evaluate our method in the popular public dataset ICDAR2013~\cite{karatzas2013icdar}. The results are presented in Tab.~\ref{tab:13det}. For the ICDAR2013 dataset, VimTS outperforms previous methods in all metrics. The proposed VimTS surpasses the previous end-to-end method TransDETR~\cite{wu2022end} by $2.3\%$ on the F-measure metric, leveraging solely image-level data and VTD-368k for training. Notably, VimTS outperforms previous methods even when trained solely on image-level data, highlighting the robust generalization and effectiveness of our method. Additionally, we also evaluate our method on the challenge dataset DSText v2~\cite{wu2024dstext}. As shown in Tab.~\ref{tab:dsdet}, our method achieves $64.2\%$ on the F-measure metric, which surpasses the previous state-of-the-art method by $2.1\%$. Additionally, our method achieves $63.7\%$ on the F-measure metric in a zero-shot manner.

\textbf{Video text Spotting.} 
For task of video text spotting, we evaluate our method on ICDAR2015 and DSText v2. The results are presented in Tab.~\ref{tab:video_spotting}. In the scene text video benchmark ICDAR2015, VimTS surpasses the previous state-of-the-art method by $1.3\%$ in terms of the $ID_{F1}$ metrics, leveraging solely image-level data and VTD-368k for training. On the DSText v2 dataset, with a zero-shot manner, VimTS outperforms the previous end-to-end method TransDETR by $1.2\%$, $3.7\%$ in $ID_{F1}$, MOTP, respectively. Additionally, as shown in Tab.~\ref{tab:video_spotting}, using real video data can further enhance the performance of our method.

\begin{figure}[htbp]
    \centering
    \begin{minipage}[c]{0.48\linewidth}
    {\includegraphics[width=4.0cm,height=3.5cm]{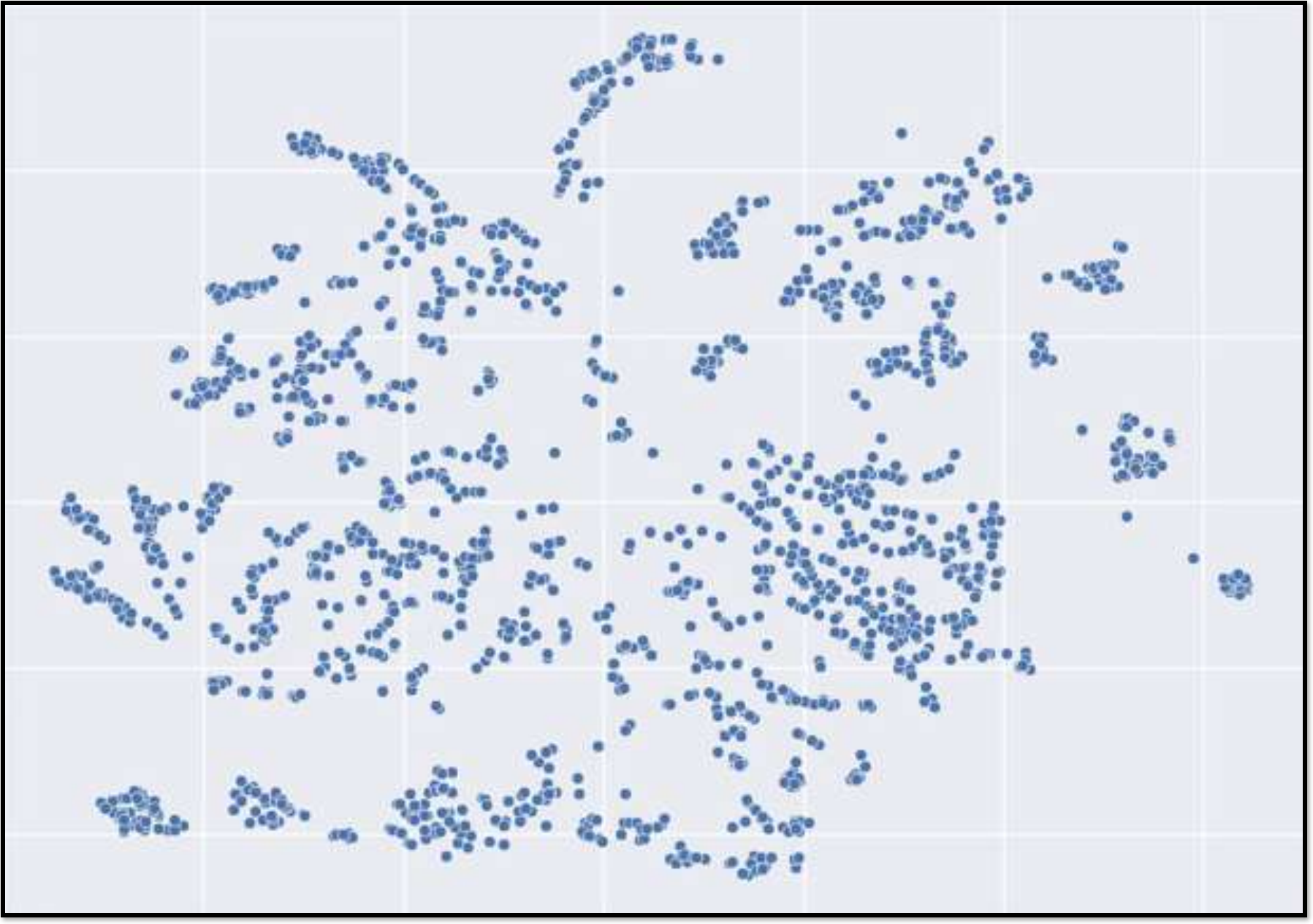}}\label{fig:vis_baseline}
    \centerline{\small{(a) Baseline}}
    \end{minipage}
    \begin{minipage}[c]{0.48\linewidth}
    {\includegraphics[width=4.0cm,height=3.5cm]{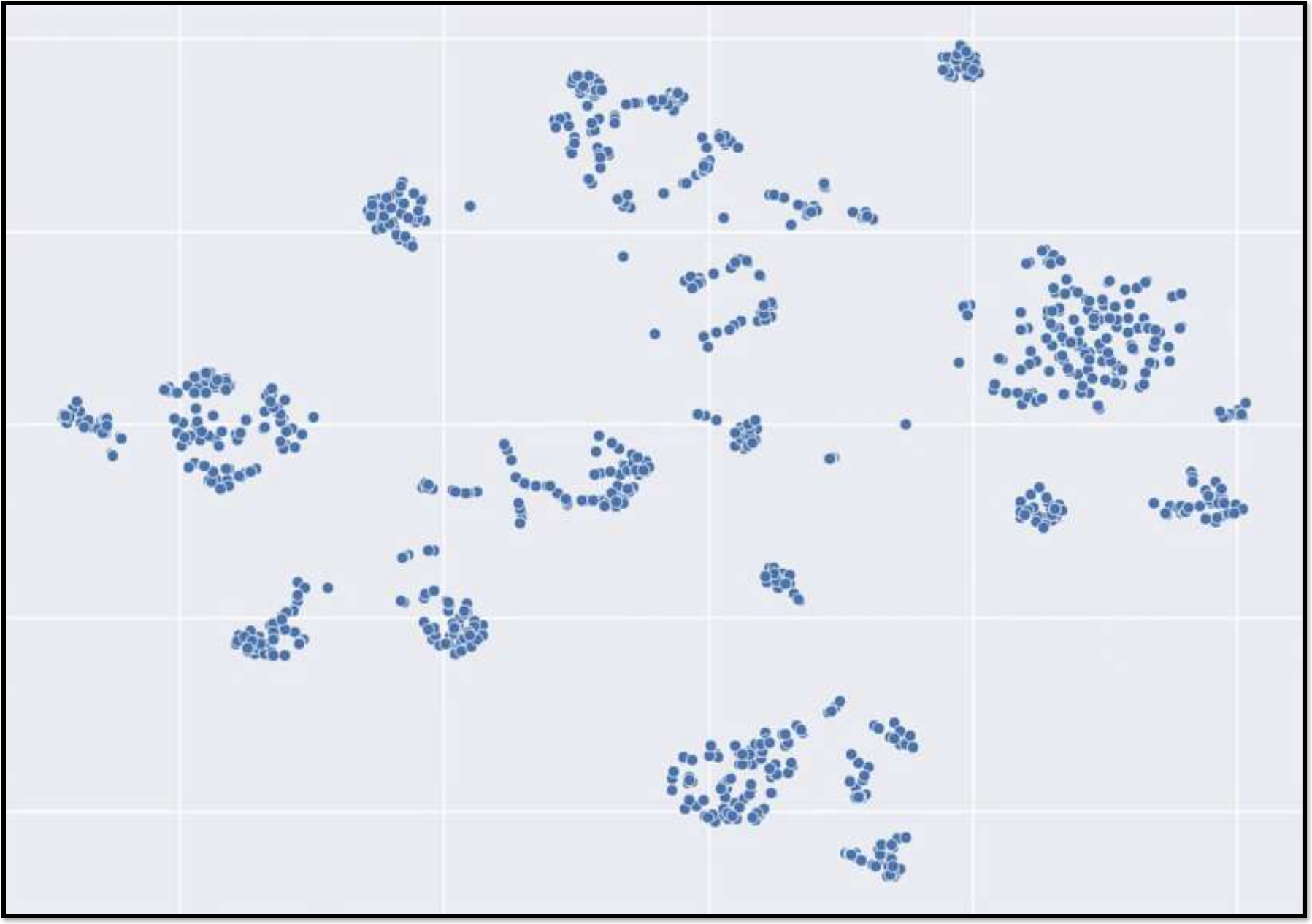}}\label{fig:vis_ours}
    \centerline{\small{(b) Ours}}
    \end{minipage}
    \caption{Distribution of the text instances in different frames via t-SNE. Best view on screen.}
    \label{fig:analysis}
\end{figure}

\subsection{Effectiveness of VTD-368k}
 
\textbf{Validating VTD-368k without real data.} 
To further evaluate the effectiveness of VTD-368k. We conduct experiments with different dataset settings on ICDAR 2015 dataset, as shown in Tab.~\ref{tab:syn_real}. For scenarios lacking real text images and text videos, we can train the model using only our synthetic data. It can be seen that even when using only synthetic data, the model is still able to perform video text spotting. The results indicate that synthetic data presents acceptable performance in scenarios where real data is not available. 

\textbf{Improving compactness by using VTD-368k.} We conducted a series of experiments on the DSText v2 dataset, employing various IoU thresholds. The ground truth and evaluation criteria were sourced from the official website. The experimental results are summarized in Tab.~\ref{tab:cp_syn_dstext}. Our observations reveal that, in the absence of VTD-368k, there is a significant decline in performance as the evaluation criteria become more stringent, specifically when the IoU threshold increases from 0.5 to 0.7. Conversely, with the inclusion of VTD-368k, the model exhibits a much smaller performance degradation under the same conditions of increasing stringency in evaluation standards. These findings suggest that VTD-368k may be able to contribute to a greater quantity of accurate output from the model.

\begin{table*}[ht]
\centering
\caption{Ablation study of using different training data on DSText v2. }
\label{tab:cp_syn_dstext}
\footnotesize 
\begin{tabular}{l|c|ccccc}
\hline
\multirow{2}{*}{Data} &\multirow{2}{*}{IOU} &  \multicolumn{5}{c}{End to End Video Text Spotting on DSText v2/\%} \\
\cline{3-7} 
& & ${
ID_{F1}}$$\uparrow$  & ${
MOTA}$$\uparrow$ & ${
MOTP}$$\uparrow$ & ${
M\mbox{-}Tracked}$$\uparrow$ & ${
M\mbox{-}Lost}$$\downarrow$ \\
\hline
image-level data & 0.5 & 12.5 & -1.0 & 82.1 & 744 & 11362 \\
image-level data + VTD-368k & 0.5 & 13.2  & -0.5  & 82.5  & 783  & 11031 \\
image-level data & 0.7 & 10.2 & -14.4 & 78.1 & 246 & 11142 \\
image-level data + VTD-368k & 0.7 &  12.3 & -1.3 & 81.9 & 758 & 10317 \\
\hline
\end{tabular}
\end{table*}

\textbf{Different baselines of using VTD-368k.} We conducted experiments to evaluate the performance of VTD-368k in comparison to other baselines. Specifically, we used TransDETR and TransVTSpotter as baseline models. The results, presented in Table~\ref{tab:syntransv1}, show that VTD-368k can improve the performance over both baselines. 
\\

\begin{table}[ht]
\centering
\caption{Ablation study of using the VTD-368k on TransDETR and TransVTSpotter (our reproduce).}
\label{tab:syntransv1}
\footnotesize 
\resizebox{.99\linewidth}{!}{%
\begin{tabular}{l|ccccc}
\hline
\multirow{2}{*}{Data} & \multicolumn{5}{c}{End to End Video Text Spotting/\%} \\
\cline{2-6} 
& ${
ID_{F1}}$$\uparrow$  & ${
MOTA}$$\uparrow$ & ${
MOTP}$$\uparrow$ & ${
M\mbox{-}Tracked}$$\uparrow$ & ${
M\mbox{-}Lost}$$\downarrow$ \\
\hline
TransDETR~\cite{wu2022end}  &  72.8 &60.9 &74.6 &33.6 &20.8 \\
 +  VTD-368k &  74.0 & 61.8 & 75.3 & 36.3 & 19.2 \\
\hline 
TransVTSpotter~\cite{wu2021bilingual}  &  61.1 & 45.8 &73.6 &29.2 & 37.2 \\ 
 +  VTD-368k &  62.5 & 46.5 & 74.2 & 30.7 & 36.5 \\
\hline
\end{tabular}
}
\end{table}

\subsection{Visualization and Analysis}
We compare our zero-shot method with two previous end-to-end approaches: TransDETR~\cite{wu2022end}, a state-of-the-art video-based method, and TESTR~\cite{zhang2022text}, a state-of-the-art image-based method, as shown in Fig.~\ref{fig:vis_results}. It can be found that even in cases of reflective and blurry text, our method consistently locates, recognizes, and tracks text instances, whereas TransDETR struggles with these scenarios. Unlike TransDETR, which relies solely on detection information between adjacent frames, our method leverages both detection and contextual information to enhance detection, recognition, and tracking capabilities, thereby achieving superior performance in video text spotting. Furthermore, compared to TESTR, which employs the same image-level data for training, our method can also locate and recognize text more accurately.

\begin{figure}[h]
    \centering
    \includegraphics[width=\linewidth]{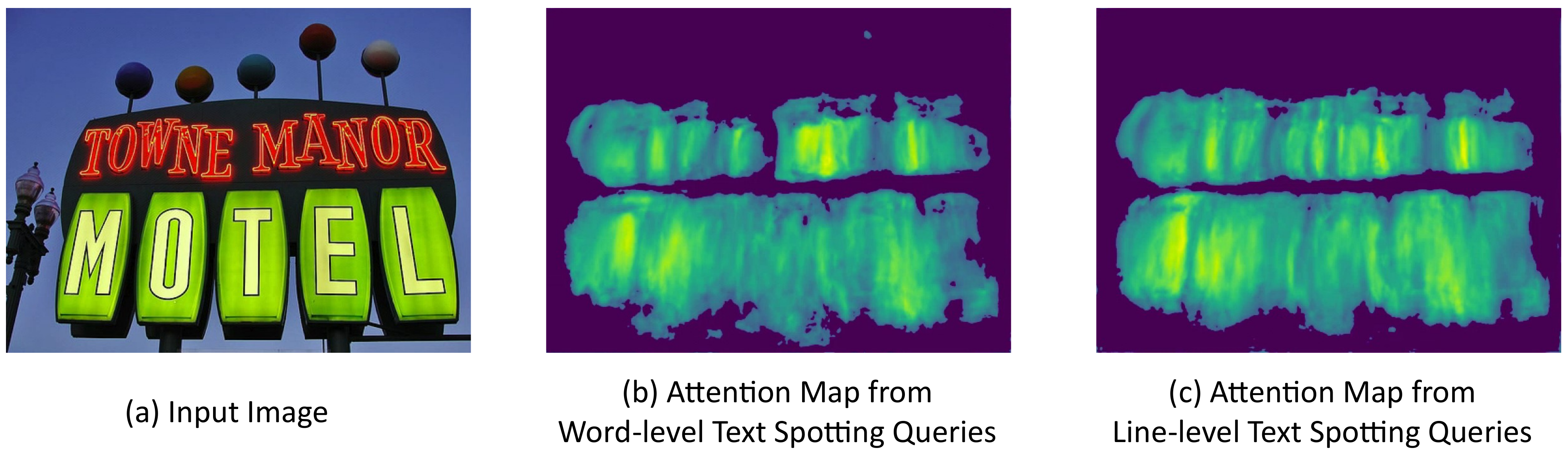}
    \caption{
    The attention map of word-level and line-level prompt queries on image features.
    }
    \label{fig:tasks_attn}
\end{figure}

To validate whether different prompt queries can learn the task-specific features, we visualize the attention map of different prompt queries on image features. When word-level queries are used, the attention map focuses on word-level features, e.g., the model treats the features of the words `TOWNE' and `MANOR' separately, as presented in Fig.~\ref{fig:tasks_attn} (b). When line-level queries are used, the  model combines the features of the words `TOWNE' and `MANOR' together, as shown in Fig.~\ref{fig:tasks_attn} (c). Intuitively, we can observe the difference between the two different queries. 

Furthermore, we also provide some qualitative results to verify why our method can achieve promising results on video text spotting by using only image-level data. To illustrate, we input the adjacent 100 frames of video into the model and visualize the distribution of the features of the text instances via t-SNE, as shown in Fig.~\ref{fig:analysis}. It can be seen that, in the proposed method, the similarity of the features of the same text in different frames is higher. Therefore, in our method, employing features from the previous frame as input queries in the current frame is effective for locating, recognizing, and tracking identical text instances even using only image-level training data.

\section{Discussion and Limitation}

Recently, Large Multi-modal Models (LMMs) have raised much attention for their robust generalization capabilities. To further show the effectiveness of our method, we conduct a cross-domain experiment on ICDAR2015 to compare with the LMMs. The evaluation process is following ~\cite{liu2023hidden}. The results are presented in Tab.~\ref{tab:mlmms}. 
The findings show the importance of developing scene text spotting methods for specific tasks. Such a specialized approach is not only more efficient in terms of requiring much fewer parameters but also necessitates significantly less training data compared to the broader application of LMMs.

\begin{figure}[!t]
    \centering
    \includegraphics[width=\linewidth]{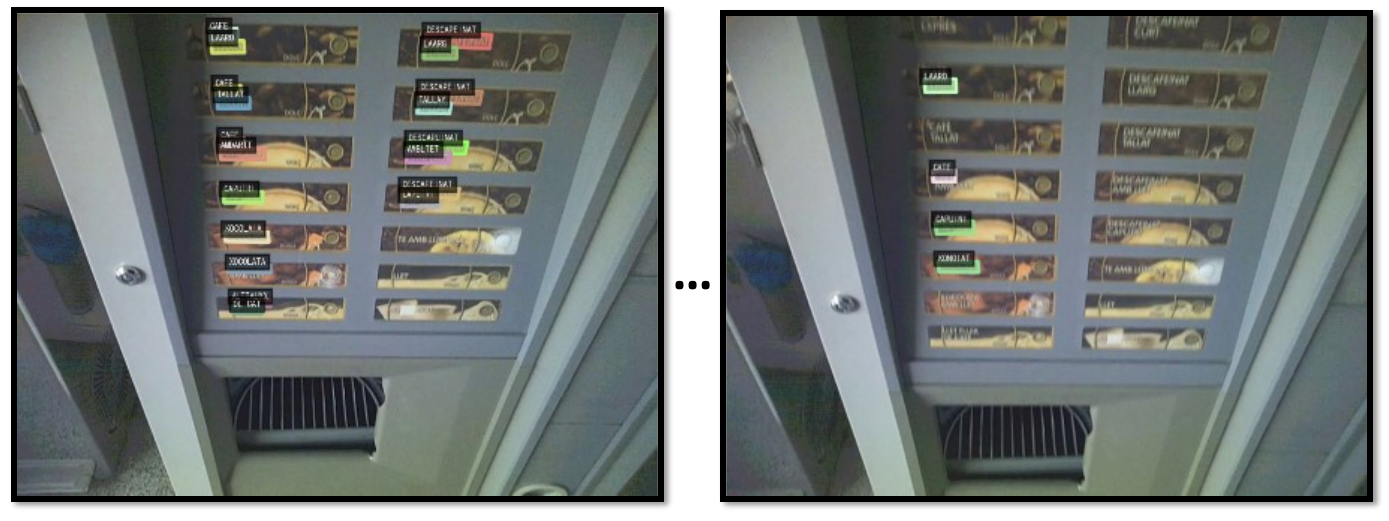}
    \caption{
        Failure cases of VimTS. When the motion amplitude is too large, the text instances are blurred seriously. Even if the text instances are correctly located and recognized in the previous frame, the model fails to locate and recognize them in the current frame.
    }
    \label{fig:error_vis}
\end{figure}

We provide some failure cases of the VimTS in Fig.~\ref{fig:error_vis}. Even when text instances in the previous frame have been successfully located and recognized, rapid movement can cause significant blurring, preventing the model from accurately locating and recognizing them in the current frame. Therefore, addressing the issue of blurred text in high-speed motion is an important direction for future exploration.

\begin{figure}[ht]
    \centering
    \centering
    \includegraphics[width=\linewidth]{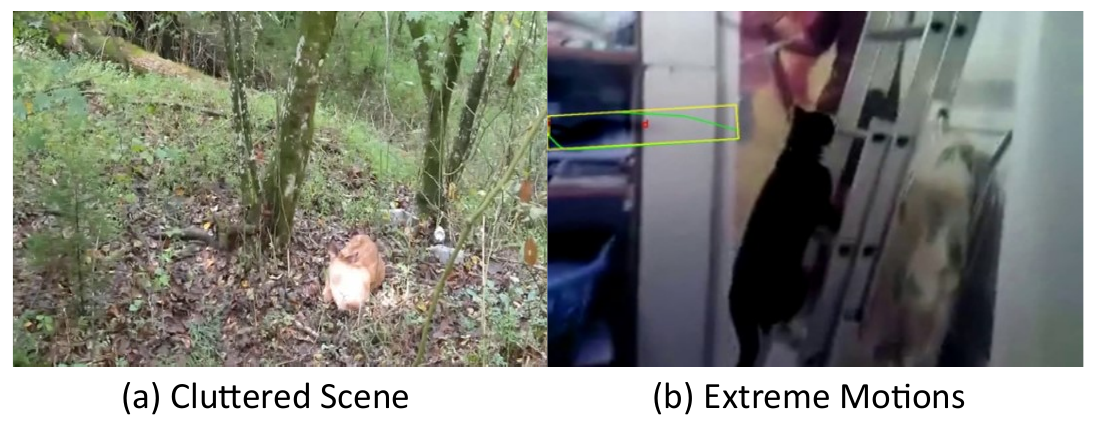}
    \caption{Failure cases of VTD-368k. (a) represents the cluttered-scene error. (b) represents the extreme motions error.}
    \label{fig:error_vtd368k}
\end{figure}

We also provide some failure synthetic samples of VTD-368k, as shown in Fig.~\ref{fig:error_vtd368k}. Errors can be observed in two scenes: cluttered scene and extreme motions scene. For the cluttered scene, due to the chaotic background, we are unable to composite the text into the video. For the extreme motions scene, the flipping or rapid movements make it difficult to establish the relationship between texts in consecutive frames during synthesis.

\begin{table}[!t]\small
\centering
\caption{Cross-domain text spotting on ICDAR2015 compared with the MLLMS. All results are tested on the `None' lexicon. OCR-related data represents the training data associated with OCR.}
\resizebox{\linewidth}{!}{
\footnotesize
\begin{tabular}{l|l | ccc}
\hline
\multirow{2}{*}{Method} & \multirow{2}{*}{OCR-related data} & \multicolumn{3}{c}{ICDAR2015/\%} \\
\cline{3-5} 
& & Precision  & Recall & F-measure \\
\hline
QWEN-VL~\cite{bai2023qwen} &  Synthetic, Natural, Doc (24.9M) & 31.9	& 42.7	& 36.5 \\
LLAVA-1.5~\cite{liu2023improved} & Natural (0.05M) & 17.8	& 8.6 & 11.6 \\
mPLUG-Owl2~\cite{ye2023mplug} & Natural (0.1M) & 17.6 & 25.6 & 20.9 \\
Monkey~\cite{li2023monkey} & Natural, Doc, Table (0.16M) & 72.3 & 29.2 & 41.6 \\
TextMonkey~\cite{liu2024textmonkey} & Natural, Doc, Table (0.41M) & 58.4 & 35.4 & 44.1 \\
InternLM-XComposer2~\cite{internlmxcomposer2} & Natural, Doc, Table (1M) & 58.9 & 36.5 & 45.1 \\
Gemini Pro~\cite{team2023gemini} & Unknown & 46.4 & 62.7 & 53.4 \\
QWEN-VL MAX~\cite{bai2023qwen} & Unknown & 50.1 & 60.4 & 54.8 \\
GPT4V~\cite{shi2023exploring} & Unknown & 57.4 & 64.6 & 60.8 \\
\hline
VimTS (Ours) & Synthetic, Natural (0.15M)& \textbf{74.3}  & \textbf{62.8} & \textbf{68.0} \\
\hline
\end{tabular}}
\label{tab:mlmms}
\end{table}

\section{Conclusion}
In this paper, we present VimTS, which improves cross-domain text spotting by enhancing the synergy among hierarchical tasks, including word-level, line-level, and video-level spotting. VimTS boosts model generalization by jointly optimizing tasks across diverse scenarios. Extensive experiments on multiple cross-domain benchmarks consistently show that our method outperforms state-of-the-art approaches by significant margins. Notably, we demonstrate that text spotters trained on still images can be effectively transferred to video text spotting, offering a promising solution given the lower annotation effort required for still images. Additionally, we highlight that current large multimodal models still face challenges in cross-domain text spotting, and addressing blurred text in high-speed motion remains a crucial area for future research.

\section*{Acknowledgement}
This research was in part supported by National Key R\&D Program of China (No.\  2022ZD0118700, 2022YFC2305100), NSFC (No.\  62225603, 62206104, 61936003), Zhuhai Industry Core and Key Technology Research Project (No.\  2220004002350). 

{
\bibliographystyle{ieeetr}
 \bibliography{arxiv}
}

\vfill 
\end{document}